# Depthwise-Dilated Convolutional Adapters for Medical Object Tracking and Segmentation Using the Segment Anything Model 2

Guoping Xu[1], Christopher Kabat[1], You Zhang[1#]

[1]The Medical Artificial Intelligence and Automation (MAIA) Laboratory

Department of Radiation Oncology

University of Texas Southwestern Medical Center, Dallas, TX 75390, USA

[#]Corresponding Author Email: You.Zhang@UTSouthwestern.edu

## Abstract

Recent advances in medical image segmentation have been driven by deep learning; however, most existing methods remain limited by modality-specific designs and exhibit poor adaptability to dynamic medical imaging scenarios. The Segment Anything Model 2 (SAM2) and its related variants, which introduce a streaming memory mechanism for real-time video segmentation, present new opportunities for prompt-based, generalizable solutions. Nevertheless, adapting these models to medical video scenarios typically requires large-scale datasets for retraining or transfer learning, leading to high computational costs and the risk of catastrophic forgetting. To address these challenges, we propose DD-SAM2, an efficient adaptation framework for SAM2 that incorporates a Depthwise-Dilated Adapter (DD-Adapter) to enhance multi-scale feature extraction with minimal parameter overhead. This design enables effective fine-tuning of SAM2 on medical videos with limited training data. Unlike existing adapter-based methods focused solely on static images, DD-SAM2 fully exploits SAM2's streaming memory for medical video objects tracking and segmentation. Comprehensive evaluations on TrackRad2025 (tumor segmentation) and EchoNet-Dynamic (left ventricle tracking) datasets demonstrate superior performance, achieving Dice scores of 0.93±0.04 and 0.97±0.01, respectively. To the best of our knowledge, this work provides an initial attempt at systematically exploring adapter-based SAM2 fine-tuning for medical video segmentation and tracking. Code, datasets, and models will be publicly available https://github.com/apple1986/DD-SAM2.

## 1. Introduction

Medical image segmentation has seen significant advancements with the rise of deep learning techniques [1]. Most existing methods focus on segmenting specific anatomical structures from static 2D images [2-4] or 3D volumetric data [5, 6]. While these approaches have demonstrated strong performance in task-specific scenarios, they exhibit critical limitations. In particular, their generalizability across different modalities is limited—models trained on one modality typically experience performance degradation when applied to another, often requiring retraining on the new dataset [7]. Moreover, these methods generally treat each image or volume independently, failing to exploit temporal information, which is critical for time-series applications such as object tracking [8].

The introduction of the Segment Anything Model (SAM) [9], trained on a large-scale dataset, has attracted significant attention to the prompt-based segmentation paradigm for its remarkable generalization capability and segmentation performance on natural images. Building upon this success, several adaptations have been proposed for medical image segmentation, including MedSAM [10] for 2D images, as well as SAM-Med3D [11] for volumetric data. More recently, SAM2 [12] was introduced, extending SAM by incorporating a streaming memory mechanism to enable real-time segmentation for both static images and videos in natural image domains. Following this advancement, a series of studies have explored fine-tuning SAM2 for medical video segmentation tasks, such as Medical SAM2 [13] and MedSAM2 [14]. These SAM-based adaptations demonstrate the effectiveness of leveraging pre-trained SAM models in medical imaging tasks, showing promising generalization and performance when fine-tuned on medical datasets. However, these approaches typically require full fine-tuning on large-scale, curated medical image datasets to effectively distill medical-domain knowledge into the SAM-based framework—a process that is both time-consuming and computationally intensive. Furthermore, extensive fine-tuning poses the risk of catastrophic forgetting, potentially compromising the model's original generalization capabilities [15-17].

Recent studies have explored parameter-efficient transfer learning (PETL) techniques, such as Adapters [18] and low-rank adaptation (LoRA) [19], in the context of SAM-based models for medical image segmentation. These approaches aim to transfer domain-specific knowledge while preserving the generalization capabilities of the original pre-trained model. The core principle of PETL is to insert lightweight modules, such as adapters, into the original architecture and update only a small subset of parameters during fine-tuning, while keeping most of the pre-trained model frozen. For instance, Medical SAM Adapter (Med-SA) [20] builds upon a standard adapter structure (see Figure 1(a)) and introduces two specialized modules—the Space-Depth Transpose-based Adapter and the Hyper-Prompting Adapter—into the image encoder and prompt encoder of SAM. These components facilitate the adaptation of 2D SAM to 3D medical images while enabling prompt-conditioned segmentation. A variant of this design, referred to as SAM-Adapter [21], retains the core structure but introduces minor modifications—most notably replacing the ReLU activation with the Gaussian error linear unit (GELU) activation. SAM-Med2D [22] proposes a novel adapter module that captures both channel-wise and spatial information, which is inserted into the image encoder of the original SAM (see Figure 1(b)). MA-SAM [23] integrates a series of standard adapters with a 3D convolution layer into the transformer blocks of SAM's image encoder to extract contextual information from volumetric medical data (see Figure 1(c)). Similarly, to further enhance the use of 3D spatial features, 3DSAM-Adapter [24] extends the standard adapter design by appending a depth-wise 3D convolutional layer, thereby improving feature extraction in 3D space (see Figure 1(d)).

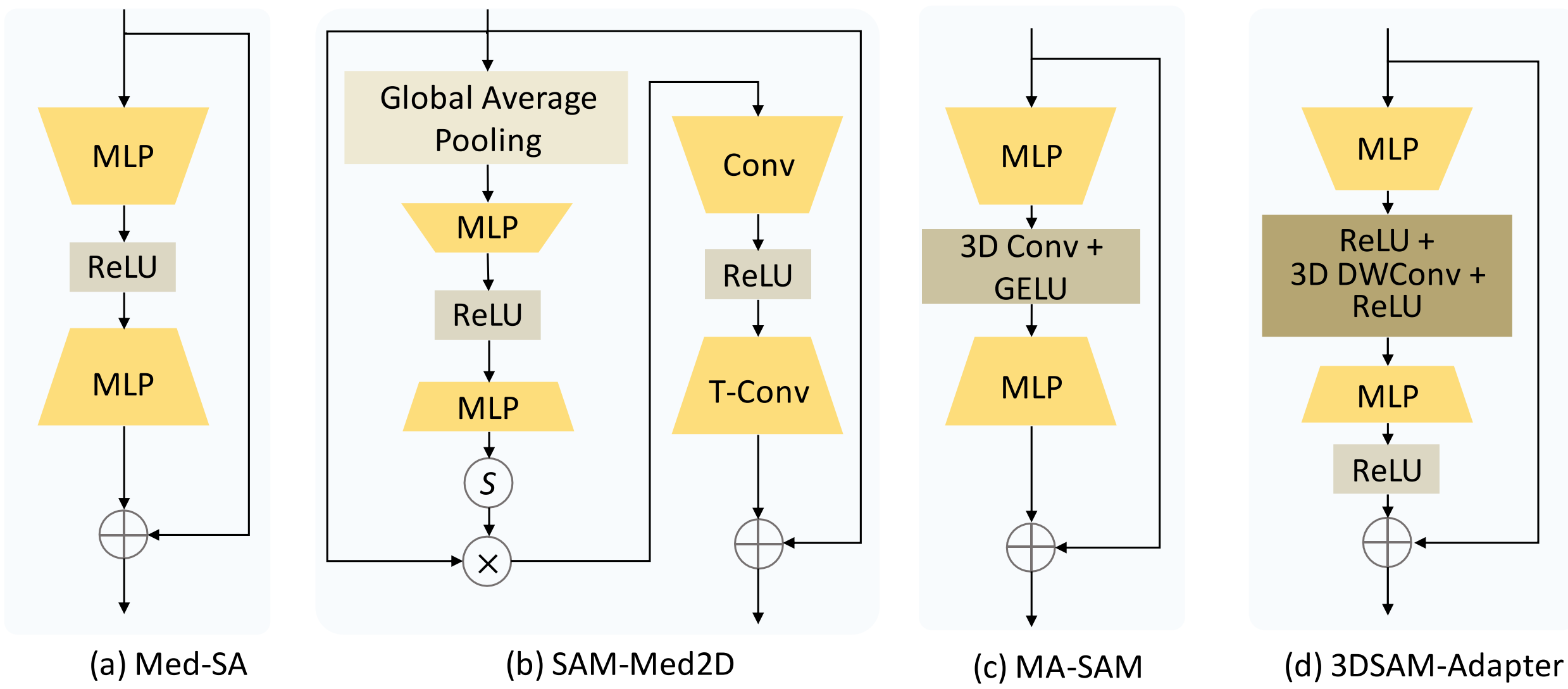

**Figure 1**. Examples of adapter architectures for fine-tuning SAM. (a) A standard adapter: the first MLP compresses the input SAM embeddings to a lower-dimensional space, and the second MLP restores them to the original dimension. (b) An adapter that encodes feature embeddings along both the channel and spatial dimensions: the left module extracts channel-wise weights, while the right module fine-tunes spatial features. (c) A 3D convolutional layer is appended to a standard adapter to capture volumetric contextual information. (d) A depth-wise 3D convolution is integrated into a standard adapter to enhance the extraction of 3D spatial features. Note that here we omit the normalization layer for simplicity.

While existing adapters have shown promise in fine-tuning SAM, two key challenges remain unaddressed. First, current methods do not explicitly incorporate multi-scale local features, which are crucial for accurately segmenting small and fine-grained anatomical structures such as tumors and vessels [14, 25]. Second, these adapter designs have been primarily developed and evaluated for static image segmentation using SAM, and their effectiveness in the context of SAM2—which supports both object tracking and segmentation for temporal image data—has yet to be fully explored. To address these limitations, we propose DD-SAM2, a novel framework based on SAM2 that incorporates a depthwise-dilated convolution-based adapter (DD-Adapter). The proposed adapter is designed to efficiently capture multi-scale local features, enhancing SAM2's capability for medical object tracking and segmentation.

Our main contributions are summarized as follows:

- We propose a novel Depthwise-Dilated Adapter (DD-Adapter) that efficiently captures multi-scale local features, enabling effective fine-tuning of SAM2 for improved domain adaptation in medical image analysis.
- We integrate the DD-Adapter into the SAM2 architecture and introduce a new adaptation framework, DD-SAM2, tailored for medical object tracking and segmentation.
- We validate our approach through comprehensive experiments on two publicly available datasets—TrackRad2025 and EchoNet-Dynamic—for tumor and left ventricle tracking and segmentation. Results show that DD-SAM2 consistently outperforms existing adapter-based baselines in both segmentation performance and tracking accuracy.

## 2. Related works

### 2.1 Segment Anything Model for Medical Imaging

The Segment Anything Model (SAM) was originally developed for natural image segmentation and trained on over 1 billion masks across 11 million images using a prompt-based segmentation paradigm. Its architecture comprises three main components: an image encoder, a prompt encoder, and a mask decoder. The image encoder, built on a stack of Vision Transformer (ViT) blocks, serves as the backbone for extracting image features and constitutes the most computationally intensive part of the model. The prompt encoder translates user-provided prompts into embedding representations, while the lightweight mask decoder generates segmentation masks using image and prompt embeddings. Following its success in natural image domains, a growing body of work has sought to adapt SAM for medical image segmentation. A primary direction has been on full fine-tuning approaches, such as MedSAM and SAM-Med3D, which fully fine-tune the pre-trained SAM on large-scale medical image datasets to integrate domain-specific knowledge.

### 2.2 Video object segmentation (VOS)

Compared to static image segmentation, a key challenge in VOS lies in effectively leveraging historical feature information or exemplary images (or masks) to guide accurate segmentation in the current frame. Recent research has predominantly focused on the semi-supervised setting, where an annotation is provided for the first frame, and the task is to track and segment the target throughout the remaining frames [26]. For example, XMem [27] introduces a memory potentiation mechanism inspired by the Atkinson–Shiffrin memory model, enabling rapid updates of sensory, working, and long-term memory to enhance temporal consistency. Given the strong image segmentation capabilities of SAM, a growing number of studies have integrated tracking algorithms with SAM to enhance performance in video object segmentation. For instance, the Track Anything Model [28] combines SAM with XMem in an interactive framework, boosting tracking and segmentation accuracy.

SAM2 builds upon the foundational strengths of SAM by integrating a streaming memory mechanism, enabling real-time video processing, and significantly advancing prompt-based segmentation capabilities. This memory component retains information about objects and prior user interactions, allowing the model to refine current frame predictions using contextual memory from previously observed frames. Building on this architecture, variants such as Biomedical SAM-2 [29], Surgical SAM2 [30], and Medical SAM2 [13] have fine-tuned SAM2 for medical applications and evaluated its performance on video object segmentation tasks. However, these adaptations typically require large-scale medical datasets and full model retraining, which imposes substantial demands in terms of data curation and computational resources.

### 2.3 Parameter-efficient fine-tuning

Parameter-efficient fine-tuning (PEFT) aims to adapt pre-trained models by updating only a small subset of parameters while keeping most of the model weights frozen. This approach has proven effective, particularly in the context of fine-tuning large language models [31, 32]. Representative adapter-based fine-tuning approaches for SAM include Med-SA [20], 3DSAM-adapter [24], SAM-Adapter [21], and SAM2-UNet [33], typically built with a down-projection, activation, and up-projection using MLPs or convolutions (see Figure 1). For LoRA-based strategies, SAMed [34] and SFR [35] apply low-rank adaptation to the SAM image encoder for medical image segmentation. Additionally, hybrid methods such as MediViSTA [36], MA-SAM [23], and Fast 3D SAM [37] combine both the adapter and LoRA mechanisms to achieve parameter-efficient fine-tuning. However, most of these methods have been applied only to SAM, which focus on static image segmentation.

The most relevant work to ours is SAM2-Adapter [38], which introduces a lightweight adapter module composed of two multilayer perceptrons (MLPs) with a GELU activation function (similar to Figure 1(a)). This approach has been evaluated across several domains, including camouflaged object detection, shadow detection, and polyp segmentation. However, it does not address the challenges of medical video object tracking and segmentation. Moreover, it overlooks the importance of multi-scale local feature representation, which can be critical for accurately segmenting anatomical structures.

To overcome these limitations, we propose a novel adapter design that integrates multi-scale locality into the adapter module. Our method enables effective fine-tuning of SAM2 for medical video object tracking and segmentation, even under limited annotated data conditions.

# 3. Method

## 3.1 Preliminary of the SAM2 architecture

SAM2 is a state-of-the-art interactive segmentation framework that extends the original SAM by introducing a streaming memory mechanism to effectively leverage historical context for both image and video segmentation tasks (see Figure 2). The architecture is composed of four key components: the image encoder, streaming memory, prompt encoder, and mask decoder. Among them, the image encoder is the most parameter-intensive module, built upon Hiera, a hierarchical vision transformer [39] pre-trained with Masked Autoencoding (MAE) [40], which facilitates efficient multi-scale feature extraction.

The streaming memory module utilizes self-attention and cross-attention mechanisms to incorporate temporal information from previous frames. It consists of three submodules: (1) a memory encoder, which fuses mask embeddings—derived via convolutional operations—with image features; (2) memory attention, which refines current frame embeddings using stored historical context; and (3) a memory bank, which maintains a queue of memory tokens in a First-In-First-Out (FIFO) manner to preserve temporal consistency. The prompt encoder transforms user inputs such as clicks, bounding boxes, or masks into embeddings and integrates them with image features. The mask decoder then processes the prompt- and memory-conditioned embeddings through two-way transformer blocks, composed primarily of self- and cross-attention layers, to generate accurate segmentation masks.

Previous studies have demonstrated the effectiveness of fine-tuning SAM2 using adapter modules [38], though the evaluation has been limited to static image segmentation, rather than video-based applications. Building upon this foundation, the present study aims to explore the integration of multi-scale local feature adapters into SAM2 and evaluate their impact on lesion tracking and segmentation in medical imaging.

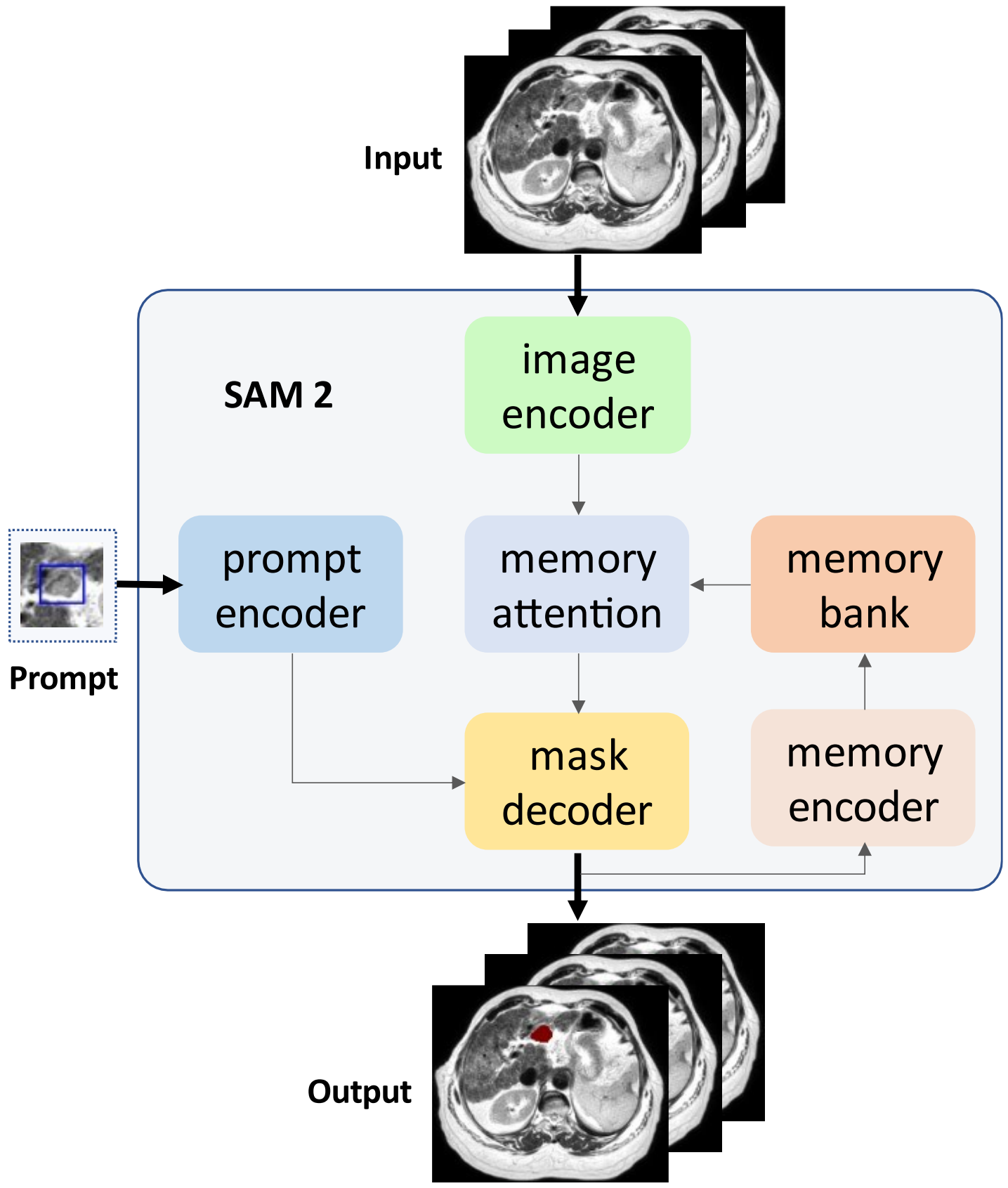

**Figure 2**. The architecture of SAM2. The model consists of four main components: the image encoder, streaming memory, prompt encoder, and mask decoder. Notably, the streaming memory module comprises three subcomponents—memory attention, memory encoder, and memory bank—designed to incorporate historical information into the feature embeddings of the current frame.

3.2 Architecture overview of DD-SAM2

The architecture of DD-SAM2 is designed to support efficient object tracking and segmentation in time-sequenced medical imaging. Built upon the standard SAM2 framework, its core innovation is the proposed depthwise-dilated convolutional adapter (DD-Adapter; see Figure 3, left). The DD-Adapter is composed of three main components: (1) two point-wise convolution (PWConv) layers for channel-wise dimensionality reduction and restoration, (2) the Gaussian Error Linear Unit (GELU) activation function, and (3) depthwise-dilated convolution (DW-DiConv) modules for multi-scale feature extraction.

Specifically, the first PWConv layer compresses the input feature maps along the channel dimension, reducing computational complexity while retaining essential information. The second PWConv layer restores the channel dimensionality to match the input, ensuring compatibility with subsequent modules. The GELU activation is employed due to its smooth non-linearity, which enables improved gradient flow and reduced activation sparsity—advantages particularly critical when fine-tuning on small medical datasets with subtle features [41]. Furthermore, GELU is used throughout the original SAM2 architecture, and we retain it for consistency. Finally, the depthwise-dilated convolution modules are incorporated to enhance spatial feature extraction. Depthwise convolutions offer a parameter-efficient alternative to standard convolutions, and by introducing dilated convolutions in parallel branches of the depthwise layers, the adapter effectively captures multi-scale local context.

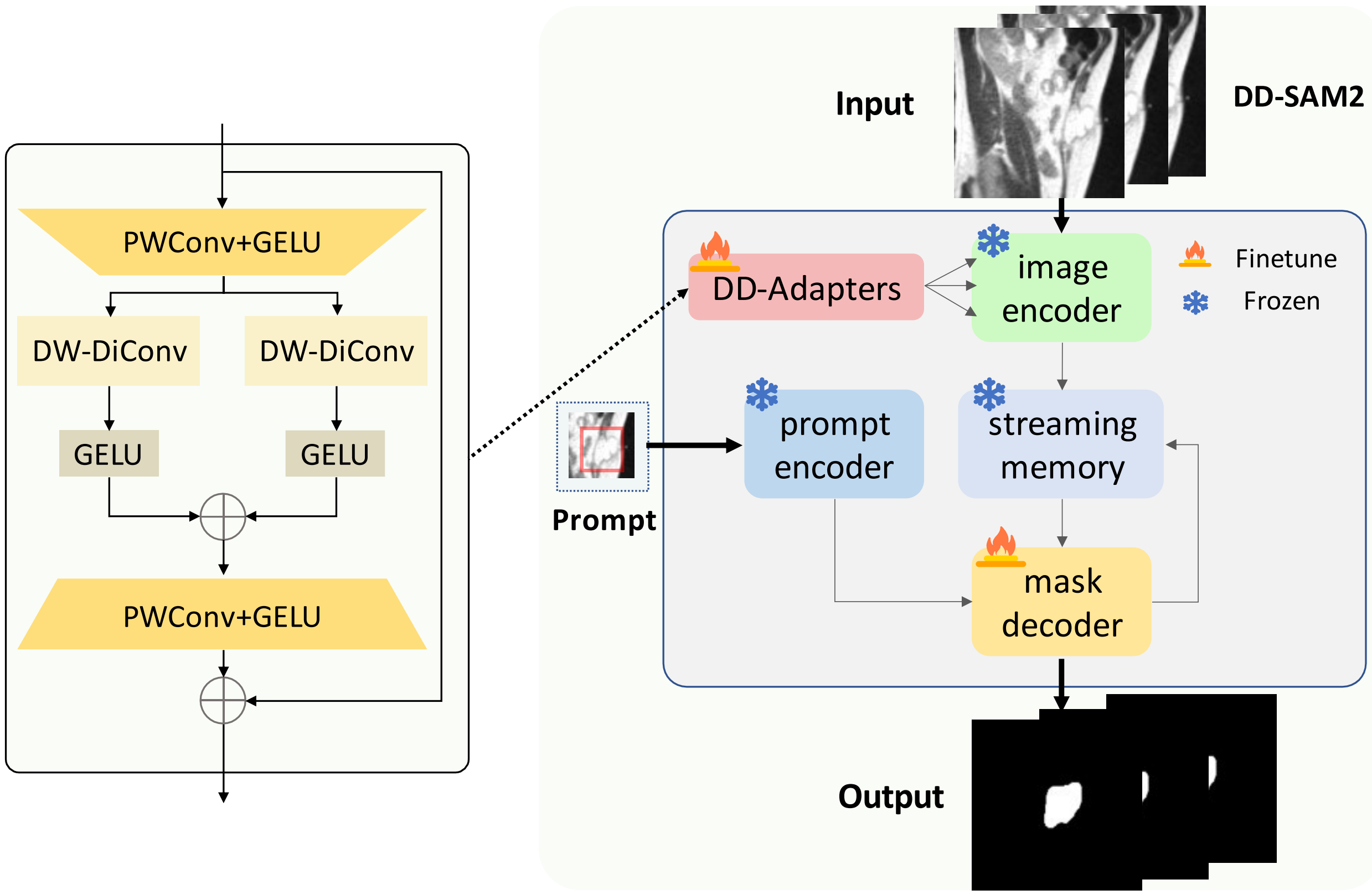

**Figure 3.** Illustration of the proposed depthwise-dilated convolutional adapter (DD-Adapter, left) and the overall DD-SAM2 framework (right), which comprises an image encoder with DD-Adapters, a prompt encoder, a streaming memory module, and a mask decoder. During fine-tuning, only the DD-Adapters and the mask decoder are updated to adapt SAM2 for medical object tracking and segmentation. The PWConv and DW-DiConv denote point-wise convolution and depthwise-dilated convolution, respectively. The GELU refers to the Gaussian Error Linear Unit, used as the activation function.

Specifically, the first PWConv layer compresses the input feature maps along the channel dimension, reducing computational complexity while retaining essential information. The second PWConv layer restores the channel dimensionality to match the input, ensuring compatibility with subsequent modules. The GELU activation is employed due to its smooth non-linearity, which enables improved gradient flow and reduced activation sparsity—advantages particularly critical when fine-tuning on small medical datasets with subtle features [41]. Furthermore, GELU is used throughout the original SAM2 architecture, and we retain it for consistency. Finally, the depthwise-dilated convolution modules are incorporated to enhance spatial feature extraction. Depthwise convolutions offer a parameter-efficient alternative to standard convolutions, and by introducing dilated convolutions in parallel branches of the depthwise layers, the adapter effectively captures multi-scale local context.

The overall processing pipeline of the proposed DD-Adapter can be conceptually divided into three stages: channel dimension reduction, multi-scale local feature extraction, and dimension recovery. This process can be mathematically formulated as follows.

First, the input feature map $f_i$ is compressed along the channel dimension using a point-wise convolution (PWConv) followed by the GELU activation function:

$$f_d = \mathrm{GELU}\big(PWConv(f_i)\big) \tag{1}$$

where $f_d$ denotes the intermediate feature map with reduced channels. This serves to reduce computational costs while preserving semantic information.

Next, $f_d$ is passed through a set of parallel depthwise-dilated convolution (DW-DiConv) branches, each with a distinct dilation rate, to extract multi-scale local features. The outputs are fused and added back to $f_d$ via residual connection:

$$f_m = f_d + \sum_{1}^{n} \mathrm{GELU}\big(DW\text{-}DiConv(f_d)\big) \tag{2}$$

where $n$ is the number of DW-DiConv branches, and each branch captures spatial patterns at different receptive fields.

Finally, the output $f_m$ is projected back to the original channel dimension using another PWConv followed by GELU activation. A residual skip connection is also added to facilitate gradient flow and feature refinement:

$$f_o = f_i + \mathrm{GELU}\big(PWConv(f_m)\big) \tag{3}$$

Here, $f_o$ denotes the final output feature map processed by the DD-Adapter. This adapter is designed to efficiently extract multi-scale contextual information, enabling effective fine-tuning of the pre-trained SAM2 model for object tracking and segmentation tasks.

### 3.3 Depthwise-dilated convolutional adapter for the image encoder

As a core component of the DD-Adapter, the structure of the depthwise-dilated convolution module is shown in Figure 4(a). We adopted depthwise convolution due to its lightweight and computationally efficient nature compared to standard convolution. To enable multi-scale local feature extraction, we incorporate dilated convolutions with varying dilation rates. For instance, a 3×3 kernel with a dilation rate of two expands the receptive field to 5×5, allowing the network to capture broader contextual information. As illustrated in Figure 4(b), we integrate the DD-Adapter after each Transformer block in the SAM2 image encoder. This design aims to inject task-specific medical domain knowledge into the pre-trained SAM2 framework, thereby enhancing its performance in object tracking and segmentation.

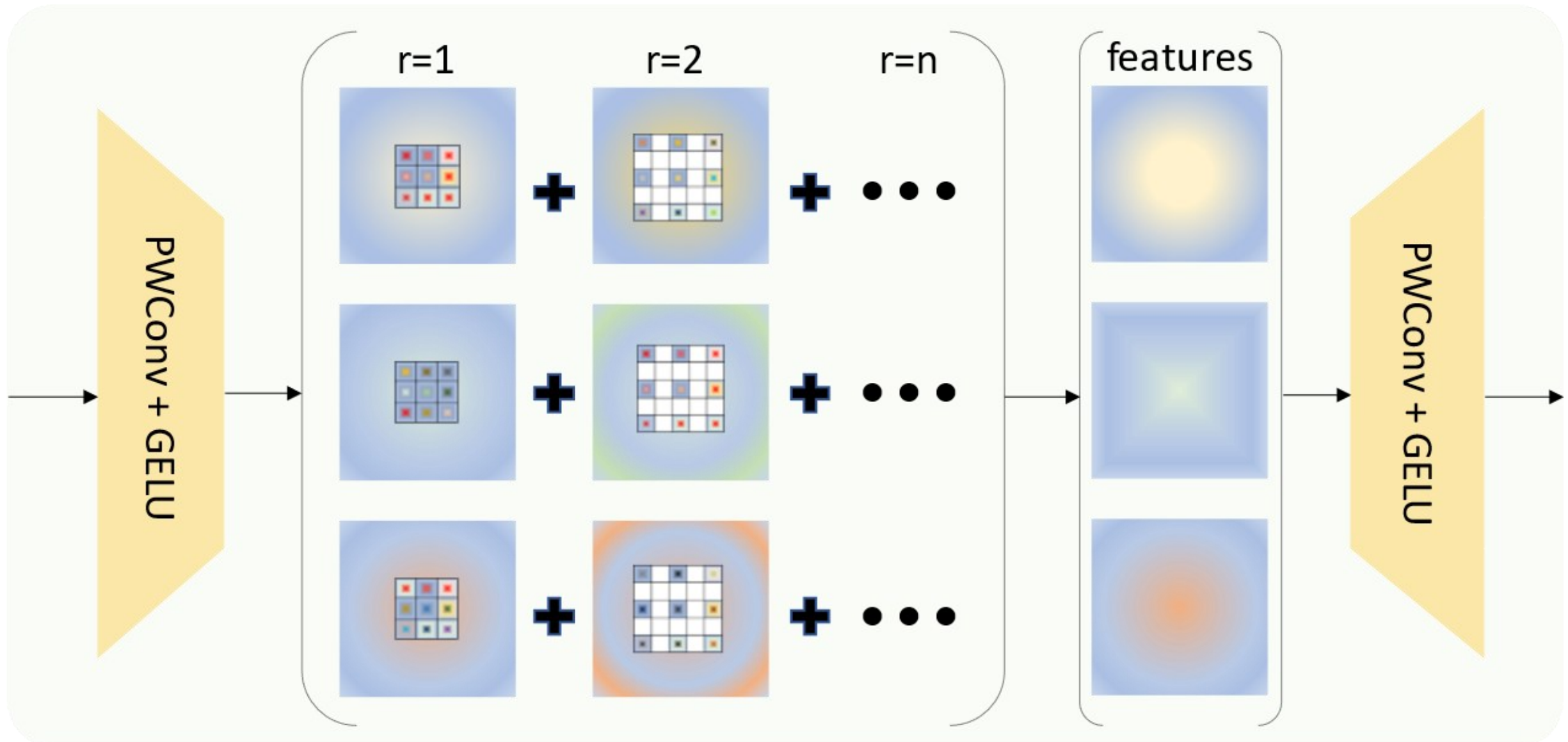

(a) The illustration of depthwise convolution in DD-Adapter.

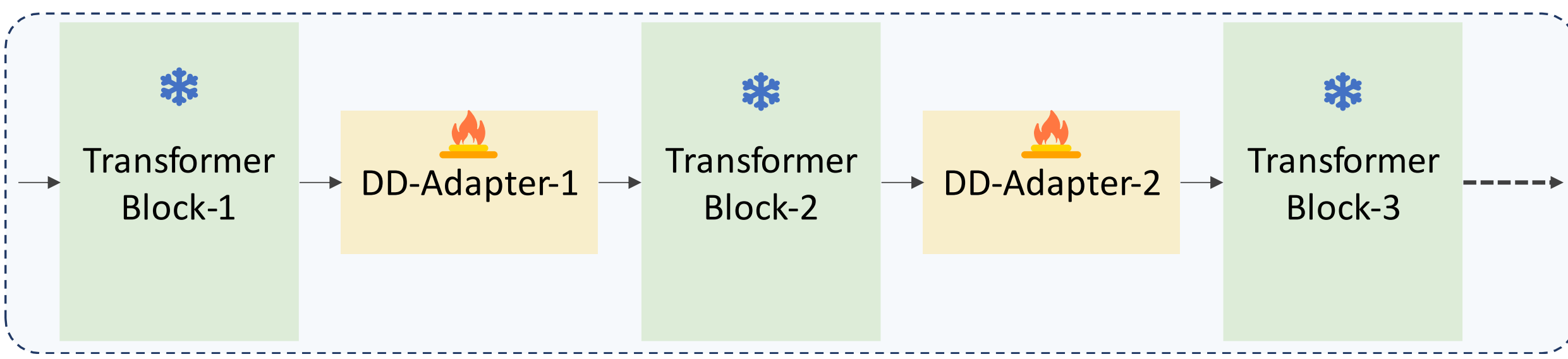

(b) The positioning of the DD-Adapter relative to the Transformer blocks of SAM2.

**Figure 4.** Top: Architecture of the proposed depthwise-dilated convolutional module (for clarity, skip connections and GELU activation functions are omitted; the *r* denotes the dilation rate of the convolution kernel). Bottom: Placement of the DD-Adapter relative to the Transformer blocks of SAM2.

# 4. Experiments

## 4.1 Dataset

To evaluate the effectiveness of DD-SAM2 in medical object tracking and segmentation, we use two publicly available datasets: TrackRAD2025[1] for tumor tracking and segmentation in MRI, and EchoNet-Dynamic [42] for left ventricle tracking and segmentation in ultrasound images.

(1) TrackRAD2025 dataset

The dataset used in this study is from the Real-time Tumor Tracking for MRI-guided Radiotherapy challenge, which focuses on tumor tracking in time-resolved sagittal 2D cine-MRI sequences [43]. It includes 477 unlabeled and 108 manually labeled cases collected from six cohorts using 0.35T and 1.5T MR-guided linear accelerators (MR-LINACs). Of these, the training set consists of 477 unlabeled and 50 labeled cases, all of which are publicly available. The remaining 58 labeled cases are reserved for final testing by the challenge organizers and are not publicly released.

[1] https://trackrad2025.grand-challenge.org/

In our experiments, we used the 50 publicly available labeled cases to evaluate the tracking and segmentation performance of our method. We found this subset sufficient to demonstrate the effectiveness of our approach. These labeled cases originate from three different cohorts, and we randomly split them into training, validation, and test sets within each cohort. In total, the split includes 34 cases for training, four for validation, and 12 for testing, as summarized in Table 1.

**Table 1.** The splitting of the 50 labeled cases from TrackRAD2025.

| Modality | Cohorts | Total Videos | Train | Val | Test |
|---|---|---|---|---|---|
| 2D Cine MRI | A-C cohort | 50 | 34 | 4 | 12 |
| | A | 25 | 17 | 2 | 6 |
| | B | 15 | 10 | 1 | 4 |
| | C | 10 | 7 | 1 | 2 |

(2) EchoNet-Dynamic dataset

The EchoNet-Dynamic dataset comprises 10,030 labeled echocardiogram videos focused on left ventricle assessment, collected from the Stanford University Hospital between 2016 and 2018. It provides a valuable benchmark for studying cardiac motion and chamber size estimation. Each video in the dataset was preprocessed by cropping and masking to remove textual annotations and content outside the scanning sector. Following the official split, the dataset comprises 7,465 training videos, 1,288 validation videos, and 1,277 testing videos (see Table 2). In this study, we used the training and validation sets to train and select the final model, respectively. All evaluations were performed on the test set.

**Table 2.** Dataset split of the EchoNet-Dynamic benchmark, showing the number of videos used for training, validation, and testing.

| Modality | Total Videos | Train | Val | Test |
|---|---|---|---|---|
| Cardiac ultrasound | 10,030 | 7,465 | 1,288 | 1,277 |

(3) CAMUS dataset

The CAMUS[2] (Cardiac Acquisitions for Multi-structure Ultrasound Segmentation) dataset is a publicly available echocardiographic dataset designed to benchmark algorithms for multi-structure cardiac segmentation from 2D ultrasound sequences [44]. It comprises 2D apical four-chamber and two-chamber view cine loops acquired from 500 patients. In this study, we utilize two-chamber view cine loops from the first 100 patients in the CAMUS dataset to evaluate the performance of our method for left ventricular tracking and segmentation across temporal sequences.

4.2 Implementation Details

All experiments were conducted on an Ubuntu system equipped with a single NVIDIA RTX 4090 GPU. We employ two versions of SAM2 as baseline models: the original pre-trained SAM2.1-Tiny (referred to as SAM2 in this study) and MedSAM2, a variant of SAM2 fine-tuned on a large-scale medical dataset containing over 455,000 3D image–mask pairs and 76,000 video frames. Across all experiments, bounding boxes are used as prompts, following the MedSAM2 protocol, as they offer a clearer and more reliable specification of target regions, particularly for organ and lesion segmentation [14]. Note that the original input spatial resolution for SAM2 is 1024×1024, whereas MedSAM2 uses 512×512. We have adopted these respective input sizes for our DD-SAM2 and DD-MedSAM2 models to ensure a fair comparison. Additionally, to simulate clinically practical deployment scenarios, only the initial bounding box from the first frame is provided as the prompt during the testing phase for all SAM-based methods. Unless stated otherwise, all experiments integrate six DD-Adapters with dilation rates of one and three into the DD-SAM2 architecture (following the first six Transformer blocks, Figure 4(b)) to achieve a balance between computational efficiency and segmentation accuracy.

[2] https://www.creatis.insa-lyon.fr/Challenge/camus/

We trained our SAM2-adapted models using an equally weighted combination of Dice loss and cross-entropy loss across both datasets. Optimization was performed using the AdamW optimizer, with an initial learning rate of 1e-4 for all adapter modules and 1e-5 for the mask decoder. For the TrackRAD2025 dataset, training was conducted over 30 epochs, with the learning rate halved at epoch 15. For the EchoNet-Dynamic dataset, to accommodate the larger training set, we fine-tuned the model for ten epochs and reduced the learning rate by half at epoch five. For each step, two videos were extracted from the TrackRAD2025 training dataset. From each video, we sampled a sub-sequence of eight video frames in length, with random beginning frames, to account for the GPU memory limit. Such randomization was performed for 64 consecutive training steps for augmentation. For the EchoNet-Dynamic dataset, per step, we extracted two videos and sampled a sub-sequence of eight frames from each video, yielding an effective batch size of two. No additional data augmentation on this dataset was performed due to the substantially larger training sample size (Table 2).

During the testing phase, the first frame of each video, along with its corresponding annotation, is provided to guide the tracking and segmentation of subsequent frames in all SAM2-based models. The target objects throughout the entire video are automatically segmented by leveraging the features from previous frames stored in the memory bank. To ensure consistency with the original input dimensions, the segmentation results are interpolated to match the original resolution. The segmentation performance is evaluated on all frames except the first, as its 'ground-truth' annotation is already provided. For quantitative evaluation, we employed the Dice Similarity Coefficient (DICE) and the Normalized Surface Distance (NSD) to assess the segmentation accuracy in terms of both volumetric overlap and local boundary alignment. In addition, the 95th percentile Hausdorff Distance (HD95) and the Average Surface Distance (ASD) were calculated to measure the surface distance errors between the predicted segmentations and the corresponding ground truth. The final reported value for each metric is the mean and standard deviation across all test videos, not across individual frames. This protocol ensures that the performance metrics reflect consistent behavior at the video level, which is more aligned with real-world clinical applications where predictions are made per video sequence rather than isolated frames.

### 4.3 Comparison with Other Methods on Object Tracking and Segmentation

We compared our frameworks with other available methods for object tracking and segmentation in medical videos, including both traditional registration-based and learning-based registration methods. For traditional registration-based methods, we employed the ANTs package[3] to perform both rigid and deformable registrations. The deformable registration is based on the symmetric image normalization method (SyN)[45]. The first frame of the video is treated as the moving image and registered to subsequent frames. The annotated mask on the first frame was propagated to other frames, serving as annotations across the series. Both the rigid and deformable registrations were based on the region-of-interest defined by the first frame's mask. Two learning-based deformable registration methods (VoxelMorph [46] and TransMorph [47]) were also evaluated. The VoxelMorph framework was implemented with both a U-Net backbone and a Swin-UNETR backbone, representing convolution-based and transformer-based architectures, respectively. In contrast, TransMorph employs a hybrid architecture that combines Transformer and convolutional components. To train both VoxelMorph and TransMorph, at each step, we randomly selected a frame from each video as the moving image and another frame as the fixed image, with the registration loss function defined on the full image frame. We trained each model for 400 epochs using an initial learning rate of 1e-4 and a batch size of 8, with the learning rate halved every 50 epochs. To ensure consistency with the SAM2-based methods, we use the same AdamW optimizer and combined Dice and cross-entropy loss functions. The best-performing model on the validation set is selected for evaluation.

[3] https://github.com/ANTsX/ANTs

In addition to the above registration-driven methods, we compared DD-SAM2 and DD-MedSAM2 with the original SAM2 and MedSAM2 frameworks. Two other memory-based tracking and segmentation methods, XMem [27] and Medical SAM2 [13], were also included for comparison.

# 5. Experiments

*5.1 Experiments for tumor tracking and segmentation on the TrackRAD2025 dataset*

(1) Comparison between different methods

The learning-based registration approaches, VoxelMorph and TransMorph, perform worse than traditional intensity-based deformable registration (Table 3). One possible cause is that the traditional intensity-based methods use the first frame and its tumor mask for registration, which provides a region-of-interest-based spatial guidance. In contrast, VoxelMorph and TransMorph solve full-sized motion fields between image pairs, with less attention on the specific tumor regions. In addition, VoxelMorph and TransMorph are trained as population-based models, while traditional methods solve motion fields via case-specific optimizations, and the latter are not susceptible to generalizability issues experienced by the former.

For memory-based tracking and segmentation methods, the segmentation performance improves substantially. Notably, DD-MedSAM2 achieves an improvement on the average Dice score of ~0.13 (in absolute terms, same in the following) over rigid registration. Furthermore, DD-MedSAM2 demonstrates an additional ~0.03 Dice gain compared to MedSAM2, demonstrating the effectiveness of our depthwise-dilated adapter design. Improvements are also observed across other key metrics, including NSD, HD95, and ASD. Similar improvements of DD-SAM2 can be observed over the original SAM2 framework. We conducted Wilcoxon signed-rank tests on all frames of the testing videos (frame-based, to improve the statistical power), which confirmed that DD-MedSAM2 consistently outperformed the other methods, with all p-values below 0.005 after Bonferroni correction, indicating statistically significant improvements.

Table 3. The quantitative results of tumor tracking and segmentation for the TrackRAD2025 dataset by different methods. Arrows are pointing in the direction of improved accuracy. Statistical tests via Wilcoxon signed-rank tests with Bonferroni correction found the differences between DD-MedSAM2 and all other methods were statistically significant ($p<0.005$).

| Method | Type | DICE↑ | NSD↑ | HD95↓[vol] | ASD↓[vol] |
|---|---|---|---|---|---|
| Registration-Rigid | Registration | 0.80±0.13 | 0.28±0.12 | 9.76±5.92 | 2.24±1.32 |
| Registration-Deformable | Registration | 0.86±0.10 | 0.33±0.08 | 5.18±2.19 | 1.46±0.55 |
| VoxelMorph-unet | Registration | 0.80±0.10 | 0.24±0.06 | 10.33±3.12 | 2.79±0.83 |
| VoxelMorph-swinunetr | Registration | 0.83±0.08 | 0.28±0.05 | 7.39±2.03 | 2.33±0.63 |
| TransMorph | Registration | 0.83±0.09 | 0.27±0.08 | 9.62±5.63 | 2.36±1.41 |
| XMem | Track & Seg | 0.89±0.11 | 0.41±0.11 | 18.67±42.43 | 2.40±4.14 |
| Medical SAM2 | Track & Seg | 0.90±0.07 | 0.39±0.11 | 4.58±3.86 | 1.21±0.81 |
| SAM2 | Track & Seg | 0.89±0.10 | 0.39±0.13 | 4.97±5.12 | 1.25±0.95 |
| DD-SAM2 | Track & Seg | 0.93±0.04 | 0.45±0.08 | 2.37±0.54 | 0.74±0.18 |
| MedSAM2 | Track & Seg | 0.90±0.05 | 0.35±0.10 | 5.85±6.71 | 1.45±1.24 |
| DD-MedSAM2 | Track & Seg | 0.93±0.04 | 0.43±0.09 | 2.81±0.89 | 0.82±0.25 |

Figure 5 presents one representative frame from three cases in the TrackRAD2025 test set, comparing segmentation results between Rigid Registration, XMem, SAM2, MedSAM2, and our proposed DD-SAM2 and DD-MedSAM2 methods (See Supplementary Materials for sequence tracking and segmentation results in video). The results demonstrate that the incorporation of DD-Adapters enhances the segmentation accuracy. XMem struggles with long video sequences, as shown by missing predictions (e.g., the first row). In contrast, SAM2-based methods—especially those enhanced with DD-Adapters—leverage a streaming memory mechanism that effectively fuses historical context, leading to more robust and consistent segmentation performance across frames.

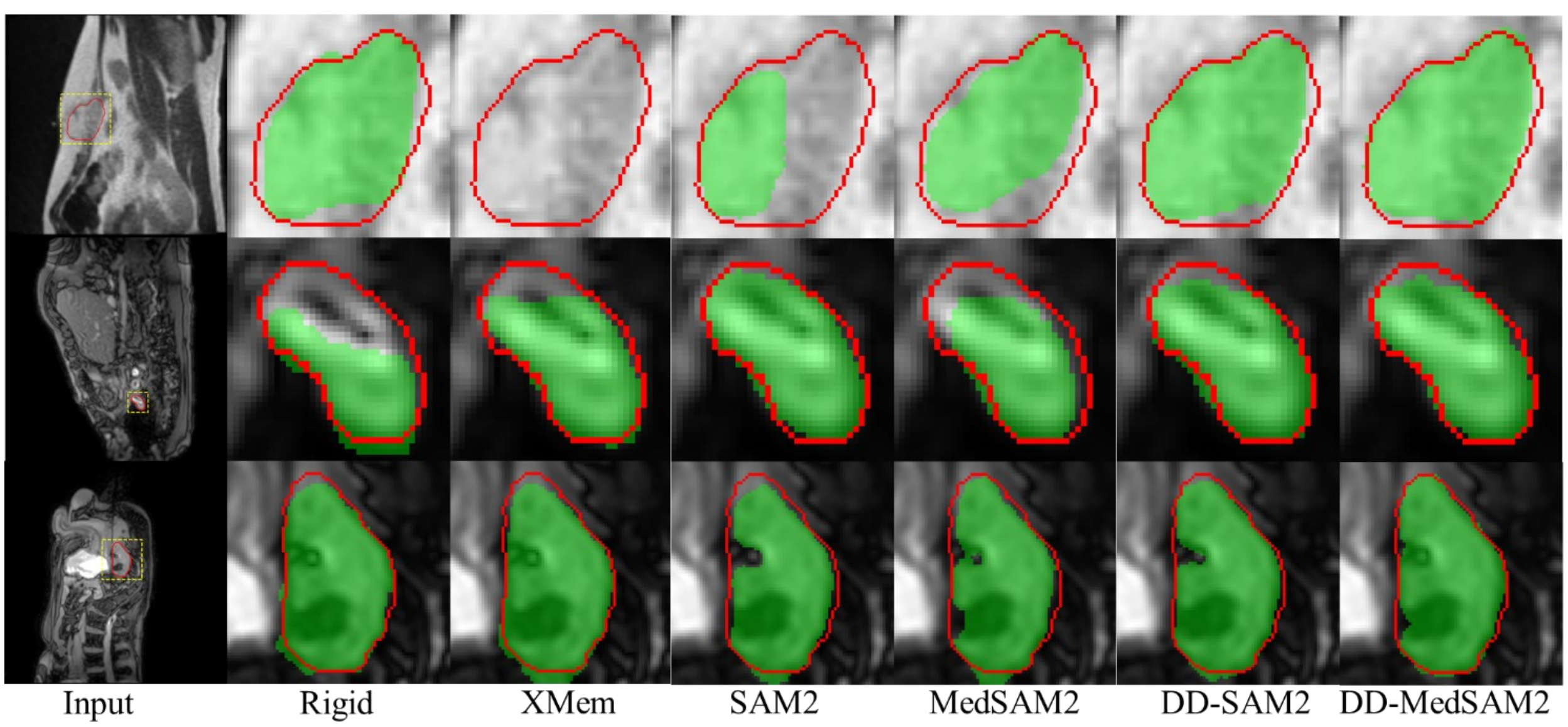

**Figure 5.** Visual comparison of segmentation results on the TrackRAD2025 dataset, between Rigid Registration, XMem, SAM2, MedSAM2, and our proposed DD-SAM2 and DD-MedSAM2 methods. The zoomed-in regions, indicated by yellow dashed boxes in the first column, show the tracked tumors. Red contours represent the 'ground-truth' boundaries, while green areas indicate the predicted segmentations by different methods.

(2) Comparison between different adapter variants for fine-tuning SAM2 and MedSAM2

As shown in Table 4, several adapter designs, along with a low-rank adaptation (LoRA) approach, were compared for fine-tuning SAM2 and MedSAM2. Our proposed DD-Adapter consistently outperforms existing adapter variants—Adp-a to Adp-d—originally introduced in Med-SAM [20], SAM-Med2D [22], MA-SAM [23] and 3DSAM-Adapter [24] (see Figure 1 for architectural details). Notably, the standard MLP-based adapter (Adp-a) achieves the lowest average Dice score (0.89), underscoring the importance of convolutional operations in capturing local structural features for medical image segmentation. Designed to extract multi-scale local features through depthwise-dilated convolutions, our DD-Adapter offers clear performance gains across all evaluation metrics—including Dice, NSD, HD95, and ASD. These improvements are consistently demonstrated in both DD-SAM2 and DD-MedSAM2, highlighting the effectiveness of the proposed architecture for adapting SAM2-based methods to medical imaging tasks.

**Table 4.** Results for using various adapters to fine-tune SAM2-based methods. Note that Adp-a/-b/-c/-d are the adapters from Med-SA, SAM-Med2D, MA-SAM, and 3DSAM-Adapter methods (see Figure 1 for the detailed structure of each). Arrows are pointing in the direction of improved accuracy.

| Method | DICE↑ | NSD↑ | HD95↓[vol] | ASD↓[vol] |
|---|---|---|---|---|
| SAM2-Adp-a | 0.89±0.04 | 0.37±0.09 | 6.29±4.13 | 1.66±1.00 |
| SAM2-Adp-b | 0.90±0.08 | 0.41±0.11 | 4.87±5.37 | 1.40±1.52 |
| SAM2-Adp-c | 0.91±0.05 | 0.40±0.09 | 3.44±1.52 | 0.99±0.35 |
| SAM2-Adp-d | 0.91±0.05 | 0.39±0.09 | 3.20±1.16 | 1.00±0.37 |
| SAM2-LoRA | 0.92±0.05 | 0.41±0.10 | 2.89±1.03 | 0.97±0.42 |
| DD-SAM2 | 0.93±0.04 | 0.45±0.08 | 2.37±0.54 | 0.74±0.18 |
| MedSAM2-Adp-a | 0.89±0.05 | 0.35±0.09 | 8.46±6.63 | 1.64±0.92 |
| MedSAM2-Adp-b | 0.92±0.05 | 0.41±0.08 | 3.63±2.83 | 0.88±0.33 |
| MedSAM2-Adp-c | 0.91±0.04 | 0.38±0.10 | 5.48±3.71 | 1.10±0.44 |
| MedSAM2-Adp-d | 0.91±0.04 | 0.39±0.09 | 5.14±3.49 | 1.06±0.43 |
| MedSAM2-LoRA | 0.91±0.06 | 0.41±0.10 | 4.75±4.01 | 0.97±0.44 |
| DD-MedSAM2 | 0.93±0.04 | 0.43±0.09 | 2.81±0.89 | 0.82±0.25 |

*5.2 Experiments for left ventricle tracking and segmentation on the EchoNet-Dynamic dataset*

Table 5 compares different methods for left ventricle tracking and segmentation on the EchoNet-Dynamic ultrasound video dataset. Our proposed method, DD-MedSAM2, outperforms all baselines by a significant margin. Specifically, it achieves a 0.06 improvement in the average DICE score and reduces the average HD95 distance error by approximately 3 voxels compared to MedSAM2, demonstrating enhanced accuracy and spatial consistency. In addition to outperforming MedSAM2, DD-MedSAM2 also shows consistent improvements across all other methods. It achieves the highest average Normalized Surface Dice (NSD) score of 0.62, more than doubling that of XMem (0.23) and significantly outperforming MedSAM2 (0.29) and SAM2 (0.27), indicating superior boundary alignment.

**Table 5.** The quantitative results of left ventricle tracking and segmentation for the EchoNet-Dynamic dataset by different methods. Arrows are pointing in the direction of improved accuracy. Statistical tests via Wilcoxon signed-rank tests with Bonferroni correction found the differences between DD-MedSAM2 and all other methods were statistically significant (p<0.005).

| Method | Type | DICE↑ | NSD↑ | HD95↓[vol] | ASD↓[vol] |
|---|---|---|---|---|---|
| Registration-Rigid | Registration | 0.86±0.05 | 0.25±0.06 | 6.27±2.35 | 1.97±0.60 |
| Registration-Deformable | Registration | 0.86±0.05 | 0.25±0.06 | 6.07±2.28 | 1.98±0.60 |
| VoxelMorph-unet | Registration | 0.84±0.06 | 0.23±0.05 | 6.40±2.19 | 2.16±0.58 |
| VoxelMorph-swinunetr | Registration | 0.71±0.14 | 0.19±0.05 | 11.19±6.07 | 2.89±0.99 |
| TransMorph | Registration | 0.86±0.04 | 0.25±0.05 | 6.11±2.50 | 1.99±0.59 |
| XMem | Track & Seg | 0.78±0.15 | 0.23±0.10 | 18.04±13.80 | 6.00±5.46 |
| Medical SAM2 | Track & Seg | 0.86±0.07 | 0.27±0.06 | 6.12±5.75 | 1.84±1.03 |
| SAM2 | Track & Seg | 0.86±0.06 | 0.27±0.05 | 6.78±3.72 | 2.07±0.82 |
| DD-SAM2 | Track & Seg | 0.96±0.04 | 0.58±0.07 | 1.72±5.10 | 0.43±0.20 |
| MedSAM2 | Track & Seg | 0.91±0.06 | 0.29±0.07 | 3.46±2.53 | 1.17±0.50 |
| DD-MedSAM2 | Track & Seg | 0.97±0.01 | 0.62±0.06 | 1.37±1.61 | 0.39±0.12 |

Figure 6 presents segmentation results from three representative cases in the EchoNet-Dynamic cardiac ultrasound dataset, comparing Rigid Registration, XMem, SAM2, MedSAM2, and our proposed DD-SAM2 and DD-MedSAM2 methods. Compared to other approaches, our methods show more stable and accurate segmentations. Rigid Registration struggles to accommodate scale and shape variations, highlighting the limitations of assuming rigid motion only. While XMem performs well on the TrackRAD2025 tumor tracking dataset, its performance substantially degrades on the EchoNet-Dynamic dataset, likely due to its inability to handle large, rapid motions in low-resolution ultrasound videos. Detailed sequence tracking and segmentation results, including figures and videos, are provided in the Supplementary Materials.

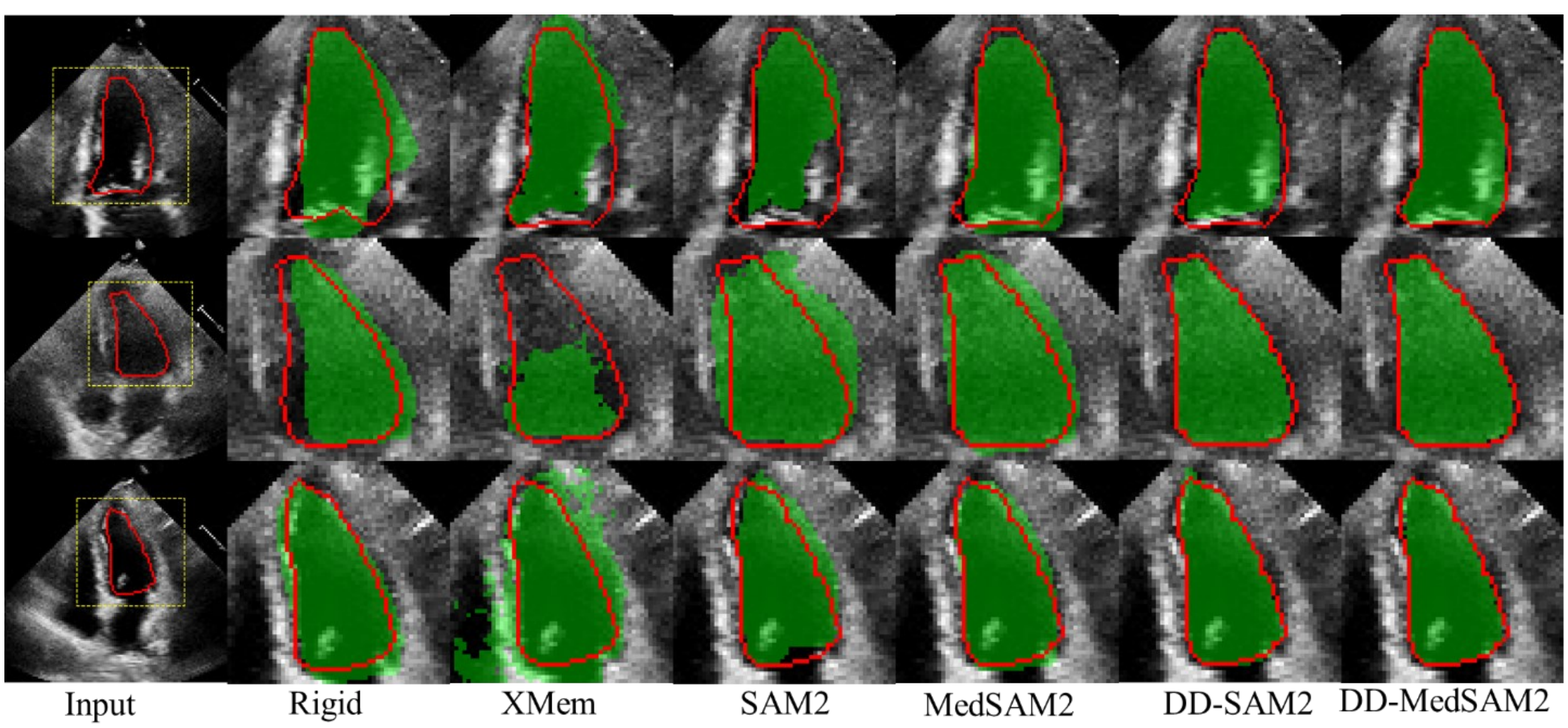

**Figure 6.** Visual comparison of segmentation results on the EchoNet-Dynamic dataset across different methods, including Rigid Registration, XMem, SAM2, MedSAM2, and our proposed DD-SAM2 and DD-MedSAM2 methods. The visualizations highlight the improved boundary accuracy achieved by models equipped with DD-Adapters.

*5.3 Additional Investigations*

(1) Generalizability evaluation

To assess the cross-domain generalization capability of our approach, we evaluated models trained exclusively on the TrackRAD2025 MRI dataset using the CAMUS ultrasound dataset, which consists of 100 patients. We employed the same inference pipeline for left ventricular tracking and segmentation across all cases. Specifically, the initial bounding box of the left ventricle—derived from the first-frame annotation—was used as the prompt input to the trained models (DD-SAM2 and DD-MedSAM2). These models then propagated prior image embeddings and predicted masks through the memory bank of the SAM2-based framework to perform tracking and segmentation in the subsequent frames.

Quantitative results are summarized in Table 6. Compared to the baseline SAM2 and MedSAM2 models, our DD-SAM2 and DD-MedSAM2 achieved improvements of approximately 0.02 in the Dice score, demonstrating enhanced performance. Notably, DD-MedSAM2 also achieved the lowest HD95 and ASD values, further confirming its generalizability to the ultrasound domain.

**Table 6.** Quantitative results of left ventricle tracking and segmentation on the CAMUS dataset. The arrows point to the direction of improved accuracy.

| Method | DICE↑ | NSD↑ | HD95↓[vol] | ASD↓[vol] |
|---|---|---|---|---|
| SAM2 | 0.84±0.05 | 0.59±0.08 | 23.22±12.19 | 4.62±2.56 |
| DD-SAM2 | 0.86±0.04 | 0.62±0.07 | 14.63±6.45 | 2.82±1.23 |
| MedSAM2 | 0.86±0.08 | 0.66±0.08 | 16.64±10.42 | 2.92±2.15 |
| DD-MedSAM2 | 0.88±0.03 | 0.64±0.06 | 12.34±5.21 | 2.34±1.18 |

(2) Ablation study of DD-Adapter

We conducted ablation experiments to assess the contributions of the pointwise convolution (PW-Conv) and dilated depthwise convolution (DD-Conv) layers in the proposed DD-Adapter. The results are summarized in Table 7. To isolate the effect of DD-Conv, we replaced it with standard convolutions (Stand-Conv) using kernel sizes of 3×3 and 7×7 to match the receptive fields of our DD-Conv. While the Dice score remained stable, degradation was observed in other metrics such as NSD, HD95, and ASD, especially when DD-Conv was removed. Similarly, removing PW-Conv also led to performance drops, indicating that both components contribute to accurate segmentation and robust tracking.

**Table 7.** Ablation study on the components of the DD-Adapter. PW-Conv, DD-Conv, and Stand-Conv refer to pointwise convolution, dilated depthwise convolution, and standard convolution, respectively.

| Method | PW-Conv | DD-Conv | Stand-Conv | DICE↑ | NSD↑ | HD95↓[vol] | ASD↓[vol] |
|---|---|---|---|---|---|---|---|
| DD-SAM2 | ✓ | ✓ | × | 0.93±0.04 | 0.45±0.08 | 2.37±0.54 | 0.74±0.18 |
| | ✓ | × | ✓ | 0.93±0.04 | 0.43±0.10 | 2.84±1.02 | 0.83±0.28 |
| | ✓ | × | × | 0.92±0.05 | 0.43±0.10 | 2.75±1.06 | 0.87±0.36 |
| | × | ✓ | × | 0.90±0.04 | 0.37±0.09 | 4.64±3.49 | 1.27±0.69 |
| DD-MedSAM2 | ✓ | ✓ | × | 0.93±0.04 | 0.43±0.09 | 2.81±0.89 | 0.82±0.25 |
| | ✓ | × | ✓ | 0.92±0.04 | 0.42±0.09 | 4.64±5.38 | 1.14±0.95 |
| | ✓ | × | × | 0.91±0.05 | 0.40±0.10 | 6.75±8.43 | 1.41±1.48 |
| | × | ✓ | × | 0.89±0.05 | 0.36±0.08 | 7.39±5.92 | 1.52±0.96 |

In addition, we analyzed both the parameter count and computational complexity (measured in FLOPs) of the proposed DD-Adapter compared to its variant using standard convolutions (Stand-Conv-Adapter). Assuming an input feature map of shape 1×C×H×W, where C is the channel dimension, the theoretical calculations are summarized in Table 8**.** Under a practical configuration with an input size of 1×96×128×128, the Stand-Conv-Adapter introduces 14 times the number of parameters and the FLOPs relative to the DD-Adapter. These results demonstrate the computational efficiency of the DD-Adapter, which preserves a large receptive field while significantly reducing resource consumption, which is beneficial for real-time or resource-constrained medical applications.

**Table 8.** Comparison between DD-Adapter and a variant with standard convolutions in terms of the number of Parameters and FLOPs.

| Adapter | Parameters | FLOPs |
|---|---|---|
| DD-Adapter | $C^2 + 9C$ | $2HWC^2+18HWC$ |
| Stand-Conv-Adapter | $15.5C^2$ | $31HWC^2$ |

(3) The number of adapters in DD-SAM2

In our default configuration, six DD-Adapters are integrated into SAM2 to balance accuracy and computational efficiency. To investigate the optimal number of adapters for DD-SAM2, we conducted an ablation study on the TrackRAD2025 dataset.

As shown in Table 9, segmentation performance generally improves as the number of adapters increases from one to six across the four-evaluation metrics, although the trend is not strictly monotonic. Beyond six adapters, the model's performance degrades. These findings suggest that inserting adapters into all transformer blocks of SAM2 may introduce overfitting or unnecessary complexity and is therefore not an optimal design choice.

**Table 9.** Quantitative results from evaluating the effect of the number of DD-Adapters in SAM2 on the TrackRAD2025 dataset.

| Number of DD-Adapters | DICE↑ | NSD↑ | HD95↓[vol] | ASD↓[vol] |
|---|---|---|---|---|
| 1 | 0.91±0.05 | 0.39±0.09 | 3.59±1.63 | 1.07±0.34 |
| 2 | 0.92±0.04 | 0.40±0.10 | 3.35±1.53 | 1.02±0.41 |
| 3 | 0.91±0.05 | 0.39±0.10 | 3.27±1.22 | 1.04±0.37 |
| 4 | 0.92±0.04 | 0.42±0.08 | 2.84±0.91 | 0.86±0.25 |
| 5 | 0.93±0.03 | 0.44±0.09 | 3.32±2.88 | 0.86±0.43 |
| 6 (DD-SAM2) | 0.93±0.04 | 0.45±0.08 | 2.37±0.54 | 0.74±0.18 |
| 7 | 0.93±0.04 | 0.44±0.08 | 2.62±0.85 | 0.81±0.24 |
| 8 | 0.93±0.04 | 0.44±0.09 | 2.67±1.10 | 0.82±0.29 |
| 9 | 0.92±0.04 | 0.42±0.09 | 3.18±1.43 | 0.94±0.34 |
| 10 | 0.90±0.08 | 0.40±0.11 | 5.30±5.49 | 1.47±1.51 |
| 11 | 0.90±0.06 | 0.41±0.11 | 4.96±4.93 | 1.34±1.21 |
| 12 | 0.91±0.05 | 0.41±0.09 | 3.21±1.45 | 0.98±0.40 |

(4) The dilation rates of DD-Adapter

To assess the impact of different dilation rates of the DD-Adapter, we evaluated five configurations—(1, 2), (1, 3), (1, 4), (1, 2, 3), and (1, 2, 3, 4)—on the TrackRAD2025 dataset. As shown in Table 10, the choice of dilation rate influences segmentation performance. Notably, the (1, 3) configuration outperforms both (1, 2) and (1, 4), indicating that an appropriate increase in dilation rates can enhance multi-scale feature representation. However, the selection of dilation rates should also consider the typical scale of the tracked objects to ensure effective feature extraction. Further increasing the number of dilations, as in (1, 2, 3) and (1, 2, 3, 4), does not yield additional gains and may even degrade the model performance. We attribute this to the grid effect associated with dilated convolutions, which can introduce artifacts and noise during multi-scale fusion (see Supplementary Materials). Based on this analysis, we adopt the (1, 3) configuration in our DD-Adapters, achieving a favorable balance between feature diversity and stability.

**Table 10.** Segmentation performance comparison of DD-SAM2 on TrackRAD2025 using different dilation rates for DD-Adapters.

| Dilated rates in DD-Adapter | DICE↑ | NSD↑ | HD95↓[vol] | ASD↓[vol] |
|---|---|---|---|---|
| [1, 3] (DD-SAM2) | 0.93±0.04 | 0.45±0.08 | 2.37±0.54 | 0.74±0.18 |
| [1, 2] | 0.92±0.05 | 0.42±0.09 | 3.40±2.27 | 1.00±0.54 |
| [1, 4] | 0.91±0.06 | 0.40±0.12 | 3.51±1.99 | 1.07±0.60 |
| [1, 2, 3] | 0.93±0.04 | 0.44±0.09 | 2.54±0.86 | 0.78±0.27 |
| [1, 2, 3, 4] | 0.92±0.04 | 0.41±0.11 | 3.75±3.69 | 1.05±0.69 |

(5) Model efficiency comparison

Table 11 provides a comparison of model efficiency across four methods, highlighting parameter count, floating-point operations (FLOPs), and inference speed measured in frames per second (FPS). For both parameter and FLOPs calculations, each input frame is defined with a shape of 1024×1024×1×3, representing a single frame with a spatial resolution of 1024×1024 and three color channels. FPS is evaluated on the TrackRAD2025 testing dataset, which contains 1,096 frames distributed across 12 cases, yielding an average of approximately 91 frames per video. All measurements were performed using a single NVIDIA RTX 4090 GPU.

We observe that introducing our Depthwise-Dilated Adapters (DD-Adapters) into SAM2 and MedSAM2 leads to a relatively minor increase of 0.54 million parameters. Specifically, DD-SAM2 and DD-MedSAM2 have 39.51M parameters, compared to the original 38.96M in SAM2 and MedSAM2. Despite the increase in FLOPs—3.9 GMac for DD-SAM2 (vs. SAM2) and 1 GMac for DD-MedSAM2 (vs. MedSAM2)—the models maintain high inference efficiency. The FPS drops only slightly by 3 FPS for DD-SAM2 (vs. SAM2) and 5 FPS for DD-MedSAM2 (vs. MedSAM2), which can be attributed to the lightweight nature of the depth-wise convolution design in our adapter module. This demonstrates that our proposed DD-Adapters can enhance model capability while incurring minimal computational overhead, making them practical for high-resolution and real-time video segmentation tasks.

**Table 11.** Model efficiency comparison in terms of parameters, floating-point operations (GMac: Giga Multiply-Accumulate operations), and inference speed (FPS: frames per second). FPS is measured on the TrackRAD2025 testing dataset.

| Method | Input size | Parameters (M) | FLOPs (GMac) | FPS |
|---|---|---|---|---|
| SAM2 | 1024×1024×1×3 | 38.96 | 113.1 | 55 |
| DD-SAM2 | 1024×1024×1×3 | 39.51 | 117.0 | 52 |
| MedSAM2 | 512×512×1×3 | 38.96 | 30.1 | 72 |
| DD-MedSAM2 | 512×512×1×3 | 39.51 | 31.1 | 67 |

Furthermore, we conducted a series of experiments to evaluate the robustness and generalizability of the proposed method under various conditions. Specifically, we analyzed the impact of different prompt types, the model's sensitivity to prompt quality, the effect of combining multiple prompt types, the influence of video length (short vs. long), and the performance under limited annotation scenarios. Detailed results and analysis are provided in the supplementary materials.

## 5. Discussion

In this study, we confirmed the efficacy of fine-tuning SAM2 with the proposed adapters for object tracking and segmentation. The proposed method demonstrates strong clinical relevance, particularly in the context of radiotherapy, where precise and real-time tumor localization is essential for accurate dose delivery and treatment safety. The ability to segment and track anatomical structures or lesions across temporally dynamic imaging sequences supports adaptive treatment planning, minimizes radiation exposure to surrounding healthy tissue, and ensures consistency across treatment sessions. Notably, our proposed DD-Adapter enables efficient fine-tuning of foundation segmentation models with a small number of additional parameters. This lightweight design facilitates rapid deployment in real-time clinical workflows, offering a practical solution for efficient and accurate tracking in radiotherapy. However, there are still several limitations that deserve exploration in the future.

(1) Integrating Prior Knowledge in Adapter Fine-Tuning

Currently, adapter-based fine-tuning strategies—typically employing bottleneck structures composed of two MLP or convolutional layers—are widely used in adapting SAM-based models to domain-specific tasks. These adapters serve to transfer knowledge from specific medical imaging domains into the pre-trained SAM model by injecting fine-tuned features during the training process. From this perspective, adapters can be interpreted as modules that learn residual features to adjust the distribution of the original representations.

While this approach has demonstrated effectiveness, the process of explicitly guiding feature learning during fine-tuning remains underexplored. A promising direction involves incorporating anatomical prior knowledge—such as organ shapes, sizes, spatial locations, and temporal relationships—into the adapter design. This integration could offer a more structured and biologically informed method for guiding feature extraction and improving segmentation accuracy. For instance, motion trajectory predictions could be leveraged to improve object localization in the tracking process by narrowing the search space based on expected anatomical movement patterns for SAM2 [48].

(2) Enhancing SAM2 Object Tracking via Memory Updates

Effectively leveraging historical information is crucial for accurate object tracking. Several approaches address this by incorporating additional temporal modules to model historical features and assist in current object tracking and segmentation tasks [48, 49]. Compared to these methods, enhancing the streaming attention module within SAM2 offers a more efficient alternative. For example, Medical SAM2 [13] introduces a self-sorting memory bank mechanism that dynamically selects the most informative historical embeddings based on confidence and dissimilarity scores. Similarly, SurgicalSAM2 [30] implements a frame pruning strategy to retain only the most relevant frames in the memory bank, thereby reducing memory usage and computational overhead.

In this study, we freeze the parameters of the streaming memory module and focus on fine-tuning the adapters within the image encoder and mask decoder. In future work, we aim to investigate memory update mechanisms in conjunction with adapter-based fine-tuning, with the goal of improving both object localization accuracy and final segmentation performance in tracking tasks.

(3) Semi-Supervised Object Tracking and Segmentation with SAM2

In this study, we focus on fine-tuning SAM2 using a proposed adapter strategy to enable robust object tracking and segmentation in time-sequenced medical images. A key limitation of our current DD-SAM2 approach lies in its dependence on fully annotated datasets for training. However, acquiring high-quality annotations for medical video sequences is both labor-intensive and costly.

A promising future direction involves extending SAM2's capabilities within a semi-supervised learning framework. Our previous work, SAMatch [50], has demonstrated the effectiveness of leveraging SAM for semi-supervised segmentation in static medical imaging. Given the vast availability of unlabeled medical video data, we plan to explore methods that harness such data to improve the fine-tuning process for SAM2 in temporal settings. This approach could substantially reduce annotation requirements while enhancing tracking and segmentation performance across time-series data.

# 6. Conclusion

In this study, we proposed a depthwise-dilated convolutional adapter (DD-Adapter), a novel design that integrates multi-scale local features within the transformer blocks for effective fine-tuning of SAM2-based methods. The correspondingly adapted models, DD-SAM2 and DD-MedSAM2, achieve both high accuracy and efficiency, highlighting their potential for real-time anatomy and tumor tracking in clinical applications to guide treatments and interventions.

# Acknowledgement

The study was supported by the US National Institutes of Health (R01 CA240808, R01 CA258987, R01 EB034691, and R01 CA280135).

# Reference

1 LeCun, Y., Bengio, Y., and Hinton, G.: 'Deep learning', Nature, 2015, 521, (7553), pp. 436-444

2 Ronneberger, O., Fischer, P., and Brox, T.: 'U-net: Convolutional networks for biomedical image segmentation', in Editor (Ed.)^(Eds.): 'Book U-net: Convolutional networks for biomedical image segmentation' (Springer, 2015, edn.), pp. 234-241

3 Zhou, Z., Siddiquee, M.M.R., Tajbakhsh, N., and Liang, J.: 'UNet++: Redesigning Skip Connections to Exploit Multiscale Features in Image Segmentation', IEEE Trans Med Imaging, 2020, 39, (6), pp. 1856-1867

4 Chen, J., Mei, J., Li, X., Lu, Y., Yu, Q., Wei, Q., Luo, X., Xie, Y., Adeli, E., Wang, Y., Lungren, M.P., Zhang, S., Xing, L., Lu, L., Yuille, A., and Zhou, Y.: 'TransUNet: Rethinking the U-Net architecture design for medical image segmentation through the lens of transformers', Med Image Anal, 2024, 97, pp. 103280

5 Milletari, F., Navab, N., and Ahmadi, S.-A.: 'V-net: Fully convolutional neural networks for volumetric medical image segmentation', in Editor (Ed.)^(Eds.): 'Book V-net: Fully convolutional neural networks for volumetric medical image segmentation' (Ieee, 2016, edn.), pp. 565-571

6 Hatamizadeh, A., Nath, V., Tang, Y., Yang, D., Roth, H.R., and Xu, D.: 'Swin unetr: Swin transformers for semantic segmentation of brain tumors in mri images', in Editor (Ed.)^(Eds.): 'Book Swin unetr: Swin transformers for semantic segmentation of brain tumors in mri images' (Springer, 2021, edn.), pp. 272-284

7 Ouyang, C., Chen, C., Li, S., Li, Z., Qin, C., Bai, W., and Rueckert, D.: 'Causality-inspired single-source domain generalization for medical image segmentation', IEEE Transactions on Medical Imaging, 2022, 42, (4), pp. 1095-1106

8 Lombardo, E., Dhont, J., Page, D., Garibaldi, C., Kunzel, L.A., Hurkmans, C., Tijssen, R.H.N., Paganelli, C., Liu, P.Z.Y., Keall, P.J., Riboldi, M., Kurz, C., Landry, G., Cusumano, D., Fusella, M., and Placidi, L.: 'Real-time motion management in MRI-guided radiotherapy: Current status and AI-enabled prospects', Radiother Oncol, 2024, 190, pp. 109970

9 Kirillov, A., Mintun, E., Ravi, N., Mao, H., Rolland, C., Gustafson, L., Xiao, T., Whitehead, S., Berg, A.C., and Lo, W.-Y.: 'Segment anything', in Editor (Ed.)^(Eds.): 'Book Segment anything' (2023, edn.), pp. 4015-4026

10 Ma, J., He, Y., Li, F., Han, L., You, C., and Wang, B.: 'Segment anything in medical images', Nat Commun, 2024, 15, (1), pp. 654

11 Wang, H., Guo, S., Ye, J., Deng, Z., Cheng, J., Li, T., Chen, J., Su, Y., Huang, Z., and Shen, Y.: 'Sam-med3d: towards general-purpose segmentation models for volumetric medical images', arXiv preprint arXiv:2310.15161, 2023

12 Ravi, N., Gabeur, V., Hu, Y.-T., Hu, R., Ryali, C., Ma, T., Khedr, H., Rädle, R., Rolland, C., and Gustafson, L.: 'Sam 2: Segment anything in images and videos', arXiv preprint arXiv:2408.00714, 2024

13 Zhu, J., Qi, Y., and Wu, J.: 'Medical sam 2: Segment medical images as video via segment anything model 2', arXiv preprint arXiv:2408.00874, 2024

14 Ma, J., Yang, Z., Kim, S., Chen, B., Baharoon, M., Fallahpour, A., Asakereh, R., Lyu, H., and Wang, B.: 'MedSAM2: Segment Anything in 3D Medical Images and Videos', arXiv preprint arXiv:2504.03600, 2025

15 Ruder, S.: 'Neural transfer learning for natural language processing', NUI Galway, 2019

16 Zhou, K.Y., Yang, J.K., Loy, C.C., and Liu, Z.W.: 'Learning to Prompt for Vision-Language Models', International Journal of Computer Vision, 2022, 130, (9), pp. 2337-2348

17 Yi, H., Qin, Z., Lao, Q., Xu, W., Jiang, Z., Wang, D., Zhang, S., and Li, K.: 'Towards general purpose medical ai: Continual learning medical foundation model', arXiv preprint arXiv:2303.06580, 2023

18 Houlsby, N., Giurgiu, A., Jastrzebski, S., Morrone, B., De Laroussilhe, Q., Gesmundo, A., Attariyan, M., and Gelly, S.: 'Parameter-efficient transfer learning for NLP', in Editor (Ed.)^(Eds.): 'Book Parameter-efficient transfer learning for NLP' (PMLR, 2019, edn.), pp. 2790-2799

19 Hu, E.J., Shen, Y., Wallis, P., Allen-Zhu, Z., Li, Y., Wang, S., Wang, L., and Chen, W.: 'Lora: Low-rank

adaptation of large language models', ICLR, 2022, 1, (2), pp. 3
20 Wu, J., Wang, Z., Hong, M., Ji, W., Fu, H., Xu, Y., Xu, M., and Jin, Y.: 'Medical SAM adapter: Adapting segment anything model for medical image segmentation', Med Image Anal, 2025, 102, pp. 103547
21 Chen, T., Zhu, L., Deng, C., Cao, R., Wang, Y., Zhang, S., Li, Z., Sun, L., Zang, Y., and Mao, P.: 'Sam-adapter: Adapting segment anything in underperformed scenes', in Editor (Ed.)^(Eds.): 'Book Sam-adapter: Adapting segment anything in underperformed scenes' (2023, edn.), pp. 3367-3375
22 Cheng, J., Ye, J., Deng, Z., Chen, J., Li, T., Wang, H., Su, Y., Huang, Z., Chen, J., and Jiang, L.: 'Sam-med2d', arXiv preprint arXiv:2308.16184, 2023
23 Chen, C., Miao, J., Wu, D., Zhong, A., Yan, Z., Kim, S., Hu, J., Liu, Z., Sun, L., Li, X., Liu, T., Heng, P.A., and Li, Q.: 'MA-SAM: Modality-agnostic SAM adaptation for 3D medical image segmentation', Med Image Anal, 2024, 98, pp. 103310
24 Gong, S., Zhong, Y., Ma, W., Li, J., Wang, Z., Zhang, J., Heng, P.A., and Dou, Q.: '3DSAM-adapter: Holistic adaptation of SAM from 2D to 3D for promptable tumor segmentation', Med Image Anal, 2024, 98, pp. 103324
25 Huang, Y., Yang, X., Liu, L., Zhou, H., Chang, A., Zhou, X., Chen, R., Yu, J., Chen, J., Chen, C., Liu, S., Chi, H., Hu, X., Yue, K., Li, L., Grau, V., Fan, D.P., Dong, F., and Ni, D.: 'Segment anything model for medical images?', Med Image Anal, 2024, 92, pp. 103061
26 Yao, R., Lin, G.S., Xia, S.X., Zhao, J.Q., and Zhou, Y.: 'Video Object Segmentation and Tracking: A Survey', Acm Transactions on Intelligent Systems and Technology, 2020, 11, (4), pp. 1-47
27 Cheng, H.K., and Schwing, A.G.: 'Xmem: Long-term video object segmentation with an atkinson-shiffrin memory model', in Editor (Ed.)^(Eds.): 'Book Xmem: Long-term video object segmentation with an atkinson-shiffrin memory model' (Springer, 2022, edn.), pp. 640-658
28 Yang, J., Gao, M., Li, Z., Gao, S., Wang, F., and Zheng, F.: 'Track anything: Segment anything meets videos', arXiv preprint arXiv:2304.11968, 2023
29 Yan, Z., Sun, W., Zhou, R., Yuan, Z., Zhang, K., Li, Y., Liu, T., Li, Q., Li, X., and He, L.: 'Biomedical sam 2: Segment anything in biomedical images and videos', arXiv preprint arXiv:2408.03286, 2024
30 Liu, H., Zhang, E., Wu, J., Hong, M., and Jin, Y.: 'Surgical sam 2: Real-time segment anything in surgical video by efficient frame pruning', arXiv preprint arXiv:2408.07931, 2024
31 Ding, N., Qin, Y.J., Yang, G., Wei, F.C., Yang, Z.H., Su, Y.S., Hu, S.D., Chen, Y.L., Chan, C.M., Chen, W.Z., Yi, J., Zhao, W.L., Wang, X.Z., Liu, Z.Y., Zheng, H.T., Chen, J.F., Liu, Y., Tang, J., Li, J.Z., and Sun, M.S.: 'Parameter-efficient fine-tuning of large-scale pre-trained language models', Nature Machine Intelligence, 2023, 5, (3), pp. 220-+
32 Fu, Z., Yang, H., So, A.M.-C., Lam, W., Bing, L., and Collier, N.: 'On the effectiveness of parameter-efficient fine-tuning', in Editor (Ed.)^(Eds.): 'Book On the effectiveness of parameter-efficient fine-tuning' (2023, edn.), pp. 12799-12807
33 Xiong, X., Wu, Z., Tan, S., Li, W., Tang, F., Chen, Y., Li, S., Ma, J., and Li, G.: 'Sam2-unet: Segment anything 2 makes strong encoder for natural and medical image segmentation', arXiv preprint arXiv:2408.08870, 2024
34 Zhang, K., and Liu, D.: 'Customized segment anything model for medical image segmentation', arXiv preprint arXiv:2304.13785, 2023
35 Li, S., Qi, L., Yu, Q., Huo, J., Shi, Y., and Gao, Y.: 'Stitching, Fine-tuning, Re-training: A SAM-enabled Framework for Semi-supervised 3D Medical Image Segmentation', IEEE Trans Med Imaging, 2025, PP
36 Kim, S., Jin, P., Chen, C., Kim, K., Lyu, Z., Ren, H., Kim, S., Liu, Z., Zhong, A., Liu, T., Li, X., and Li, Q.: 'MediViSTA: Medical Video Segmentation via Temporal Fusion SAM Adaptation for Echocardiography', IEEE J Biomed Health Inform, 2025, PP
37 Shen, Y., Li, J., Shao, X., Inigo Romillo, B., Jindal, A., Dreizin, D., and Unberath, M.: 'Fastsam3d: An efficient segment anything model for 3d volumetric medical images', in Editor (Ed.)^(Eds.): 'Book Fastsam3d: An efficient segment anything model for 3d volumetric medical images' (Springer, 2024, edn.),

pp. 542-552
38 Chen, T., Lu, A., Zhu, L., Ding, C., Yu, C., Ji, D., Li, Z., Sun, L., Mao, P., and Zang, Y.: 'Sam2-adapter: Evaluating & adapting segment anything 2 in downstream tasks: Camouflage, shadow, medical image segmentation, and more', arXiv preprint arXiv:2408.04579, 2024
39 Ryali, C., Hu, Y.-T., Bolya, D., Wei, C., Fan, H., Huang, P.-Y., Aggarwal, V., Chowdhury, A., Poursaeed, O., and Hoffman, J.: 'Hiera: A hierarchical vision transformer without the bells-and-whistles', in Editor (Ed.)^(Eds.): 'Book Hiera: A hierarchical vision transformer without the bells-and-whistles' (PMLR, 2023, edn.), pp. 29441-29454
40 He, K., Chen, X., Xie, S., Li, Y., Dollár, P., and Girshick, R.: 'Masked autoencoders are scalable vision learners', in Editor (Ed.)^(Eds.): 'Book Masked autoencoders are scalable vision learners' (2022, edn.), pp. 16000-16009
41 Hendrycks, D., and Gimpel, K.: 'Gaussian error linear units (gelus)', arXiv preprint arXiv:1606.08415, 2016
42 Ouyang, D., He, B., Ghorbani, A., Yuan, N., Ebinger, J., Langlotz, C.P., Heidenreich, P.A., Harrington, R.A., Liang, D.H., Ashley, E.A., and Zou, J.Y.: 'Video-based AI for beat-to-beat assessment of cardiac function', Nature, 2020, 580, (7802), pp. 252-256
43 Wang, Y., Lombardo, E., Thummerer, A., Blocker, T., Fan, Y., Zhao, Y., Papadopoulou, C.I., Hurkmans, C., Tijssen, R.H.N., Gorts, P.A.W., Tetar, S.U., Cusumano, D., Intven, M.P., Borman, P., Riboldi, M., Dudas, D., Byrne, H., Placidi, L., Fusella, M., Jameson, M., Palacios, M., Cobussen, P., Finazzi, T., Haasbeek, C.J.A., Keall, P., Kurz, C., Landry, G., and Maspero, M.: 'TrackRAD2025 challenge dataset: real-time tumor tracking for MRI-guided radiotherapy', Med Phys, 2025, 52, (7), pp. e17964
44 Leclerc, S., Smistad, E., Pedrosa, J., Ostvik, A., Cervenansky, F., Espinosa, F., Espeland, T., Berg, E.A.R., Jodoin, P.M., Grenier, T., Lartizien, C., Dhooge, J., Lovstakken, L., and Bernard, O.: 'Deep Learning for Segmentation Using an Open Large-Scale Dataset in 2D Echocardiography', IEEE Trans Med Imaging, 2019, 38, (9), pp. 2198-2210
45 Avants, B.B., Epstein, C.L., Grossman, M., and Gee, J.C.: 'Symmetric diffeomorphic image registration with cross-correlation: evaluating automated labeling of elderly and neurodegenerative brain', Med Image Anal, 2008, 12, (1), pp. 26-41
46 Balakrishnan, G., Zhao, A., Sabuncu, M.R., Guttag, J., and Dalca, A.V.: 'VoxelMorph: A Learning Framework for Deformable Medical Image Registration', IEEE Trans Med Imaging, 2019, 38, (8), pp. 1788-1800
47 Chen, J., Frey, E.C., He, Y., Segars, W.P., Li, Y., and Du, Y.: 'TransMorph: Transformer for unsupervised medical image registration', Med Image Anal, 2022, 82, pp. 102615
48 Yang, C.-Y., Huang, H.-W., Chai, W., Jiang, Z., and Hwang, J.-N.: 'Samurai: Adapting segment anything model for zero-shot visual tracking with motion-aware memory', arXiv preprint arXiv:2411.11922, 2024
49 Cuttano, C., Trivigno, G., Rosi, G., Masone, C., and Averta, G.: 'SAMWISE: Infusing wisdom in SAM2 for Text-Driven Video Segmentation', arXiv preprint arXiv:2411.17646, 2024
50 Xu, G., Qian, X., Shao, H.C., Luo, J., Lu, W., and Zhang, Y.: 'A segment anything model-guided and match-based semi-supervised segmentation framework for medical imaging', Med Phys, 2025, 52, (6), pp. 4513-4527

# Supplementary information

## 1. Impact of Different Prompt Types on DD-MedSAM2

We evaluated different prompt types, including bounding box, mask, and sparse points (five randomly selected points per object), on the DD-MedSAM2 framework. The results are presented in Table 1. Among these results, using the 'ground-truth' mask from the first frame as the prompt achieved the best segmentation performance across all metrics. However, this approach requires significantly more manual annotation effort compared to others. Notably, the bounding box prompt yielded better performance than the point-based prompt, resulting in a 0.09 improvement in Dice score. This suggests that even coarse spatial priors, such as bounding boxes, provide more robust guidance to the model than sparsely sampled point cues.

Table 1. Comparison of DD-MedSAM2's performance across different prompt types. Arrows indicate the direction of improved accuracy, consistent with the subsequent tables.

| Types | DICE↑ | NSD↑ | HD95↓[vol] | ASD↓[vol] |
|---|---|---|---|---|
| Box | 0.93±0.04 | 0.43±0.09 | 2.81±0.89 | 0.82±0.25 |
| Mask | 0.93±0.04 | 0.46±0.08 | 2.65±0.91 | 0.73±0.22 |
| Points | 0.84±0.13 | 0.35±0.12 | 15.87±15.56 | 4.58±5.64 |

## 2. Investigating DD-Adapter's Model Sensitivity to Prompt Quality

(1) *Sensitivity of varying numbers of point prompts*

To assess the impact of the number of point prompts on segmentation performance, we conducted an evaluation by randomly selecting 1 to 10 foreground points as input prompts for the DD-MedSAM2 model. The corresponding performance metrics are summarized in Table 2.

Table 2. DD-MedSAM2's performance under varying numbers of point prompts.

| Num of Points | DICE↑ | NSD↑ | HD95↓[vol] | ASD↓[vol] |
|---|---|---|---|---|
| 1 | 0.78±0.26 | 0.35±0.13 | 16.35±22.54 | 5.35±11.96 |
| 2 | 0.80±0.19 | 0.34±0.12 | 21.38±17.97 | 5.67±6.63 |
| 3 | 0.81±0.19 | 0.35±0.13 | 15.22±14.74 | 4.43±5.69 |
| 4 | 0.83±0.15 | 0.35±0.13 | 15.81±15.62 | 4.89±6.28 |
| 5 | 0.84±0.13 | 0.35±0.12 | 15.87±15.56 | 4.58±5.64 |
| 6 | 0.86±0.11 | 0.36±0.12 | 16.37±14.84 | 4.37±4.88 |
| 7 | 0.87±0.10 | 0.36±0.12 | 15.63±13.63 | 3.60±3.33 |
| 8 | 0.88±0.09 | 0.36±0.11 | 15.50±13.94 | 3.05±2.22 |
| 9 | 0.88±0.08 | 0.37±0.10 | 10.43±8.89 | 2.43±1.73 |
| 10 | 0.89±0.09 | 0.37±0.10 | 11.01±14.14 | 2.39±2.08 |

We observed an overall consistent performance improvement as positive points increased. With only one point, the Dice score was relatively low (0.78), and the model showed high variability in boundary delineation, as reflected in elevated HD95 and ASD values. As more points were introduced, especially when beyond five, the model began to stabilize and achieved better segmentation quality, both in overlap-based and boundary-based metrics.

This result highlights the benefit of incorporating multiple spatial cues when using sparse prompts, as they help constrain the model toward better localization and structure-aware segmentation, even in the absence of full object masks.

(3) *Sensitivity to object mask variations*

To investigate how the variations of the initial object mask affect the segmentation performance, we applied two morphological operations, erosion and dilation, to systematically modify the mask size. Specifically, we altered the size of the mask in the first frame using various kernel sizes for these operations and evaluated their impact on DD-MedSAM2.

For erosion operations, we observed that as the kernel size increased, the performance of DD-MedSAM2 consistently declined (Table 3). In particular, larger kernel sizes for erosion (⩾13) resulted in a significant drop in the Dice score, with degraded boundary accuracy metrics (NSD, HD95, ASD), some of which became unavailable due to highly degraded predictions. This highlights the sensitivity of the model to under-segmented prompts.

Table 3. The impacts from variations of object mask prompts with erosion operations on DD-MedSAM2.

| **Erosion kernel size** | **DICE↑** | **NSD↑** | **HD95↓[vol]** | **ASD↓[vol]** |
|---|---|---|---|---|
| 1 | 0.93±0.04 | 0.46±0.08 | 2.65±0.91 | 0.73±0.22 |
| 3 | 0.92±0.04 | 0.43±0.08 | 4.16±4.51 | 0.91±0.58 |
| 5 | 0.91±0.06 | 0.38±0.07 | 3.63±1.77 | 0.96±0.25 |
| 7 | 0.88±0.07 | 0.33±0.07 | 4.14±1.80 | 1.16±0.27 |
| 9 | 0.86±0.08 | 0.29±0.08 | 4.63±1.78 | 1.37±0.35 |
| 11 | 0.85±0.07 | 0.27±0.08 | 5.26±1.88 | 1.62±0.50 |
| 13 | 0.70±0.32 | 0.18±0.10 | N/A | N/A |
| 15 | 0.65±0.31 | 0.15±0.10 | N/A | N/A |
| 17 | 0.58±0.34 | 0.14±0.10 | N/A | N/A |
| 19 | 0.56±0.34 | 0.13±0.10 | N/A | N/A |

In contrast, the performance degradation with mask dilations was more gradual (Table 4). Even at a kernel size of 19, the Dice score remained relatively high (0.73 vs. 0.56 for dilation vs. erosion), suggesting that DD-MedSAM2 is more tolerant of over-segmented prompts than under-segmented ones. This robustness could be due to the model's ability to refine coarse predictions when given sufficient contextual information.

These findings emphasize the importance of carefully selecting the prompt. Using a mask that is too small (e.g., heavily eroded) can significantly impair the segmentation accuracy, while a moderately over-segmented mask retains more resilience in the model performance.

Table 4. The impacts from variations of object mask prompts with dilation operations on DD-MedSAM2.

| **Dilation kernel size** | **DICE↑** | **NSD↑** | **HD95↓[vol]** | **ASD↓[vol]** |
|---|---|---|---|---|
| 1 | 0.93±0.04 | 0.46±0.08 | 2.65±0.91 | 0.73±0.22 |
| 3 | 0.93±0.04 | 0.43±0.09 | 2.75±0.82 | 0.81±0.2 |
| 5 | 0.92±0.04 | 0.38±0.09 | 3.02±0.83 | 1.00±0.34 |
| 7 | 0.90±0.05 | 0.34±0.09 | 3.63±1.00 | 1.28±0.47 |
| 9 | 0.87±0.07 | 0.30±0.09 | 4.20±1.16 | 1.60±0.63 |
| 11 | 0.85±0.08 | 0.27±0.08 | 4.88±1.28 | 2.02±0.82 |
| 13 | 0.82±0.09 | 0.25±0.08 | 5.91±1.57 | 2.53±1.02 |
| 15 | 0.79±0.10 | 0.23±0.07 | 6.94±1.77 | 3.07±1.21 |
| 17 | 0.76±0.12 | 0.21±0.07 | 7.87±2.06 | 3.66±1.50 |
| 19 | 0.73±0.13 | 0.20±0.06 | 8.92±2.28 | 4.32±1.78 |

(4) *Impact of bounding box sizes*

We further evaluated the impact of varying the size of the bounding box prompts on the segmentation performance. Specifically, we randomly enlarged each side of the original bounding box (extracted from the 'ground-truth' annotation of the first frame) by up to a defined number of pixels. For instance, for a value of 11, the bounding box was randomly expanded by 0 to 11 pixels in the top, bottom, left, and right directions.

As shown in Table 5, the performance of DD-MedSAM2 gradually declined as the bounding box size increased. While moderate enlargement (up to 5–7 pixels) led to only minor performance degradations, larger expansions (e.g., 11 pixels or more) affected the boundary precision and stability more. For example, when enlarging up to 19 pixels, the Dice score dropped from 0.92 to 0.83, and the average surface distance (ASD) more than tripled.

This indicates that overly loose bounding boxes introduce more irrelevant background regions, making it more difficult for the model to accurately focus on the target structure. However, the model retains moderate robustness to bounding box variations, which supports its applicability in clinical scenarios where precise bounding box annotations may not always be available.

Table 5. The impacts from variations of bounding box prompts with random enlargement operations per side (by 1–19 pixels) on DD-MedSAM2.

| **Enlargement by pixels** | **DICE↑** | **NSD↑** | **HD95↓[vol]** | **ASD↓[vol]** |
|---|---|---|---|---|
| 1 | 0.92±0.04 | 0.43±0.09 | 2.87±0.87 | 0.84±0.25 |
| 3 | 0.92±0.05 | 0.42±0.08 | 2.93±0.88 | 0.88±0.28 |
| 5 | 0.91±0.06 | 0.41±0.09 | 3.05±0.87 | 0.96±0.32 |
| 7 | 0.91±0.06 | 0.41±0.09 | 4.15±2.85 | 1.14±0.53 |
| 9 | 0.90±0.07 | 0.40±0.09 | 3.60±1.23 | 1.14±0.46 |
| 11 | 0.89±0.09 | 0.37±0.10 | 4.69±2.76 | 1.49±0.81 |
| 13 | 0.89±0.09 | 0.37±0.10 | 4.69±2.76 | 1.49±0.81 |
| 15 | 0.87±0.09 | 0.36±0.09 | 9.40±8.04 | 2.29±1.43 |
| 17 | 0.87±0.09 | 0.36±0.09 | 9.40±8.04 | 2.29±1.43 |
| 19 | 0.83±0.13 | 0.32±0.09 | 12.19±6.75 | 3.15±1.39 |

### 3. Combinations of Different Prompt Types

To further investigate the effect of combining different prompt types, we conducted experiments using dual prompts: one derived from the annotated mask of the first frame, and the other from the middle frame of each video sequence. The first-frame prompt was a bounding box, while the middle-frame prompt varied across three types: mask, bounding box, and 10 positive points.

As summarized in Table 6, combining the first-frame box with a middle-frame mask yielded the best overall performance, with a Dice score of 0.94 and improved boundary accuracy (lowest HD95 and ASD). This demonstrates that multi-frame guidance, particularly with precise mask information, can enhance model performance by providing richer spatial and temporal context. Combining bounding boxes from the first and middle frames as prompts achieved a performance similar to the original single-prompt framework, while the use of mid-frame points as additional prompts slightly degrades the segmentation accuracy, potentially due to the difficulty in precisely representing object shape and location with sparse point prompts.

Moreover, we hypothesize that the suboptimal performance of the box + points configuration may also stem from the lack of adaptation of the prompt encoder to multi-type and multi-frame prompts, as the encoder was not retrained for the new prompts. Fine-tuning the prompt encoder to better handle such heterogeneous prompt combinations could potentially improve the model's ability to integrate and interpret spatial cues across frames.

Table 6. The impacts from combined prompt types (first frame: box, middle frame: mask, box, or points) on DD-MedSAM2.

| Prompt Type | DICE↑ | NSD↑ | HD95↓[vol] | ASD↓[vol] |
|---|---|---|---|---|
| Box (Frame 1) | 0.93±0.04 | 0.43±0.09 | 2.81±0.89 | 0.82±0.25 |
| Box (Frame 1) + Mask (Mid-Frame) | 0.94±0.04 | 0.46±0.08 | 2.27±0.72 | 0.69±0.18 |
| Box (Frame 1) + Box (Mid-Frame) | 0.93±0.04 | 0.43±0.08 | 2.81±0.97 | 0.81±0.23 |
| Box (Frame 1) + Points (Mid-Frame) | 0.91±0.07 | 0.40±0.09 | 4.83±3.72 | 1.29±0.72 |

## 4. Impacts of Video Lengths on Model Accuracy

*(1) Results by training with different lengths of video sub-sequences*

We investigated the impact of different training sub-sequence lengths on the performance of DD-MedSAM2 by retraining the model with 4, 8, and 12 consecutive frames sampled from each video. As shown in Table 7, the model trained with 8-frame sequences achieved the best overall performance, surpassing both the shorter (4-frame) and longer (12-frame) sequences across all evaluation metrics. Specifically, the 8-frame setup yielded the highest Dice score (0.93), indicating superior segmentation accuracy, along with the lowest HD95 and ASD values, reflecting more precise boundary alignment and spatial consistency.

Table 7. Performance under various lengths of training sub-sequences for DD-MedSAM2.

| Training Length | DICE↑ | NSD↑ | HD95↓[vol] | ASD↓[vol] |
|---|---|---|---|---|
| 4 | 0.90±0.09 | 0.42±0.09 | 4.66±4.42 | 0.99±0.64 |
| 8 | 0.93±0.04 | 0.43±0.09 | 2.81±0.89 | 0.82±0.25 |
| 12 | 0.88±0.12 | 0.40±0.11 | 5.43±4.95 | 1.25±1.05 |

We hypothesize that excessively short sequences may provide insufficient temporal context for robust feature propagation, while overly long sequences may introduce redundancy or temporal noise that hinders the model’s learning efficiency. Additionally, the observed performance differences may be influenced by the memory bank management strategy employed during training. In DD-MedSAM2, previous frames are stored in a First-In-First-Out (FIFO) manner. This means that using sequences that are too short may result in inadequate memory context, while sequences that are too long may include outdated or less relevant information, both of which can negatively affect the model's ability to accurately track and segment target structures in the current frame.

Following the original training protocol of SAM2, we adopted 8-frame sequences for training DD-MedSAM2, as this configuration provides a balanced temporal window that supports optimal feature representation and memory utilization.

*(2) Inference with short and long videos*

To assess the effect of temporal progression on segmentation performance, we divided each video in the testing set into two segments based on frame indices: frames 1–40 and 41 to the end. The corresponding evaluation metrics for each segment are summarized in Table 8. The results reveal that while the Dice score remains relatively stable across all segments, the performance tends to degrade in terms of NSD, HD95, and ASD as the sequence progresses. This trend suggests increasing uncertainty and segmentation challenges with longer temporal contexts. Providing prior knowledge, such as motion trajectories, or adding corrective prompts in the middle could help mitigate this degradation. Additionally, it remains crucial to explore effective methods for fusing features from previous frames to update the current feature maps for improved object recognition and segmentation.

Table 8. Segmentation performance across different temporal segments on TrackRad2025 with DD-MedSAM2.

| **Index of Frame** | **DICE↑** | **NSD↑** | **HD95↓[vol]** | **ASD↓[vol]** |
|---|---|---|---|---|
| [1, 40] | 0.923±0.046 | 0.525±0.086 | 2.527±0.836 | 0.673±0.218 |
| [41, end] | 0.923±0.049 | 0.515±0.132 | 2.795±0.968 | 0.729±0.291 |

*(3) Evaluating potential tracking drift or error accumulation under erroneous prompts*

To investigate the impact of prompt errors and the associated error accumulation, we introduced controlled degradation by shrinking the boundary of the first-frame mask by a specified number of pixels (see Figure 1). This simulates the presence of imperfect or noisy prompts that may occur during deployment. Similar to Table 8, we also separated the test set into two sub-sets to evaluate potential error accumulation. Table 9 summarizes the segmentation results under increasing levels of boundary erosion.

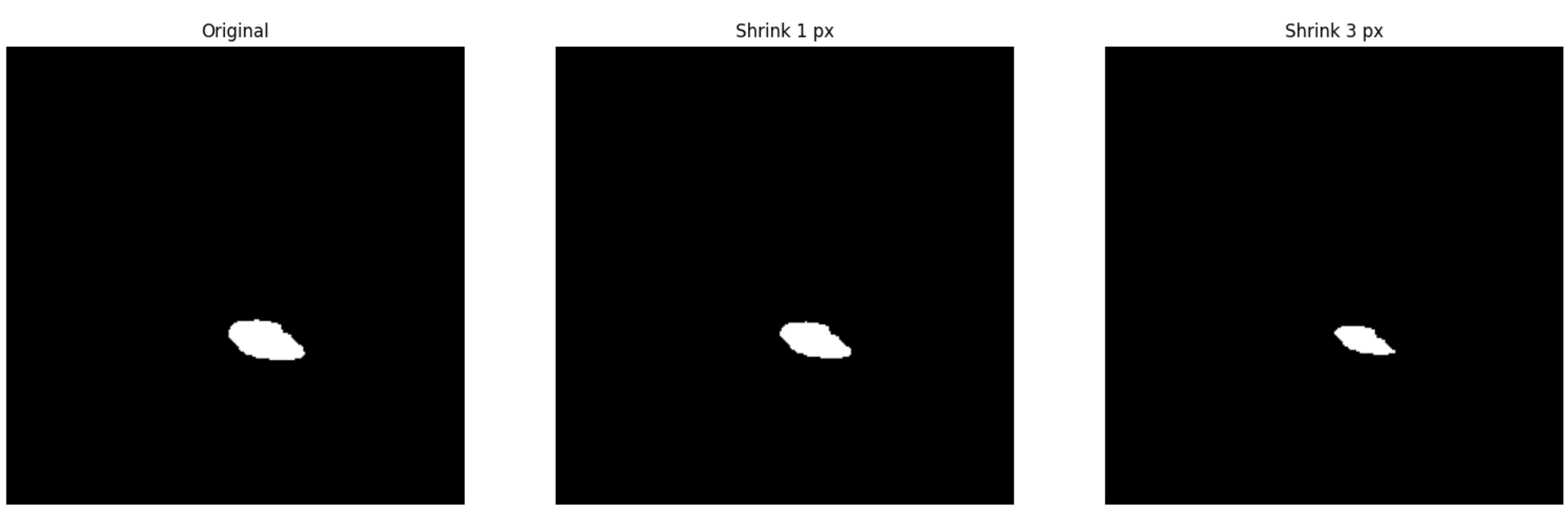

Figure 1. Illustration of boundary erosion applied to the prompt mask on the first frame. From left to right: original ‘ground-truth’ mask, mask eroded by 1 pixel (px), and mask eroded by 3 pixels. This controlled erosion simulates varying levels of prompt degradation used to assess model robustness against inaccurate inputs and to evaluate potential error accumulations.

As shown, the segmentation performance degrades as the erosion degree increases. While the Dice score remained relatively stable under mild perturbations, metrics such as NSD, HD95, and ASD began to deteriorate—especially when beyond 3 pixels of erosion. In extreme cases (e.g., 7 or 9 pixels), the segmentation quality becomes so poor that the surface-based metrics (HD95 and ASD) are no longer computable, indicated as “N/A” in the table.

Table 9. Performance degradation under increasing boundary erosions of the prompt mask. The first-frame mask is used as a prompt. Errors are introduced by shrinking the mask boundary by a specified number of pixels.

| Shrink (px) | Index of Frame | DICE↑ | NSD↑ | HD95↓[vol] | ASD↓[vol] |
|---|---|---|---|---|---|
| 0 | [1, 40] | 0.94±0.04 | 0.56±0.07 | 2.26±0.75 | 0.55±0.13 |
| 0 | [41, end] | 0.93±0.04 | 0.53±0.13 | 2.70±1.03 | 0.68±0.27 |
| 1 | [1, 40] | 0.93±0.04 | 0.52±0.07 | 2.45±0.87 | 0.58±0.15 |
| 1 | [41, end] | 0.92±0.04 | 0.52±0.13 | 4.43±5.26 | 0.86±0.72 |
| 3 | [1, 40] | 0.88±0.07 | 0.41±0.08 | 4.05±1.94 | 0.99±0.34 |
| 3 | [41, end] | 0.89±0.07 | 0.45±0.14 | 3.62±1.29 | 0.86±0.29 |
| 5 | [1, 40] | 0.83±0.07 | 0.34±0.08 | 5.20±1.73 | 1.43±0.47 |
| 5 | [41, end] | 0.87±0.06 | 0.40±0.14 | 4.58±1.71 | 1.16±0.48 |
| 7 | [1, 40] | 0.63±0.30 | 0.23±0.14 | N/A | N/A |
| 7 | [41, end] | 0.66±0.32 | 0.25±0.17 | N/A | N/A |
| 9 | [1, 40] | 0.53±0.30 | 0.19±0.13 | N/A | N/A |
| 9 | [41, end] | 0.57±0.36 | 0.22±0.17 | N/A | N/A |

Interestingly, we observed that DD-MedSAM2 demonstrates resilience to error propagation over time for some cases. For example, when the initial prompt is eroded by 5 pixels, the Dice score for the early segment ([1, 40]) is 0.83, but improves to 0.87 in the later segment ([41, end]), despite no additional correction to the prompt. This suggests that the model leverages not only the initial prompt but also the image content of the current frame and historical features from previous frames to refine its predictions. Such behavior indicates a capacity of DD-MedSAM2 to mitigate prompt degradation through temporal and spatial context aggregation. Exploring strategies to handle low-quality prompts remains a promising direction. One potential approach is to leverage large vision models, such as DINOv2 [1] or DINOv3 [2], to extract robust semantic features. These features can then be used to generate more reliable prompts, thereby improving the quality of the prompt and enhancing segmentation performance.

## 5. Performance Under Limited Training Data

To investigate model performance under limited annotations, we retrained DD-SAM2 and DD-MedSAM2 on the TrackRAD2025 dataset using only 1 to 6 training cases, respectively. The results, presented in Table 10, indicate that both models maintain competitive performance across all four evaluation metrics, even with as few as one or two training cases. This demonstrates the strong generalization capability of the pretrained SAM2 backbone when integrated with our proposed DD-Adapter for fine-tuning. As expected, increasing the number of training cases generally improved the performance, although minor fluctuations were observed. For example, DD-SAM2 trained with 3 cases performed slightly worse than with 1, 2, 4, 5, or 6 cases.

Table 10. The quantitative results of tumor tracking and segmentation for the TrackRAD2025 dataset are presented for different numbers of training cases.

| Method | Num of Training Cases | DICE↑ | NSD↑ | HD95↓[vol] | ASD↓[vol] |
|---|---|---|---|---|---|
| DD-SAM2 | 1 | 0.90±0.06 | 0.39±0.12 | 4.58±4.01 | 1.30±1.00 |
| DD-SAM2 | 2 | 0.90±0.06 | 0.39±0.12 | 4.68±4.28 | 1.30±0.96 |
| DD-SAM2 | 3 | 0.88±0.08 | 0.37±0.12 | 5.25±5.19 | 1.56±1.30 |
| DD-SAM2 | 4 | 0.90±0.07 | 0.41±0.13 | 4.53±4.01 | 1.28±1.01 |
| DD-SAM2 | 5 | 0.91±0.04 | 0.42±0.12 | 4.07±2.88 | 1.18±0.90 |
| DD-SAM2 | 6 | 0.91±0.07 | 0.41±0.13 | 3.97±3.54 | 1.15±0.87 |
| DD- MedSAM2 | 1 | 0.88±0.10 | 0.35±0.11 | 5.65±4.27 | 1.32±0.97 |
| DD- MedSAM2 | 2 | 0.89±0.06 | 0.35±0.10 | 4.79±3.49 | 1.19±0.69 |
| DD- MedSAM2 | 3 | 0.92±0.04 | 0.41±0.09 | 3.79±1.88 | 0.99±0.43 |
| DD- MedSAM2 | 4 | 0.92±0.04 | 0.40±0.07 | 4.07±2.91 | 0.90±0.29 |
| DD- MedSAM2 | 5 | 0.92±0.04 | 0.41±0.07 | 3.47±1.22 | 0.90±0.29 |
| DD- MedSAM2 | 6 | 0.92±0.04 | 0.41±0.09 | 3.26±1.12 | 0.92±0.31 |

**6. Comparison with Multi-Scale Standard Convolution Alternatives for the Adapter Structure**

To evaluate the use of standard convolution alternatives within the adapter structure, we replaced the depthwise-dilated convolution in the DD-Adapter with multi-scale variants (Stand-Conv), implemented as parallel standard convolutions with different kernel sizes. The results are summarized in Table 11. Using kernel sizes of 3×3 and 5×5, Stand-Conv underperformed the original DD-Adapter. Expanding the receptive field further with 3×3 and 7×7 convolutions did not improve MedSAM2's performance, although a slight gain was observed for SAM2. This suggests that larger convolutional kernels may be more beneficial for high-resolution inputs, as SAM2 operates on 1024×1024 images, whereas MedSAM2 uses 512×512.

Table 11. Comparison between DD-Adapter and multi-scale alternatives (Stand-Conv) using parallel standard convolutions of different kernel sizes.

| Framework | Description | DICE↑ | NSD↑ | HD95↓[vol] | ASD↓[vol] |
|---|---|---|---|---|---|
| DD-MedSAM2 | DD-Adapter | 0.93±0.04 | 0.43±0.09 | 2.81±0.89 | 0.82±0.25 |
| DD-MedSAM2 | Stand-Conv (3, 5) | 0.92±0.04 | 0.42±0.09 | 4.17±3.52 | 1.01±0.53 |
| DD-MedSAM2 | Stand-Conv (3, 7) | 0.92±0.04 | 0.42±0.09 | 4.64±5.38 | 1.14±0.95 |
| DD-SAM2 | DD-Adapter | 0.93±0.04 | 0.45±0.08 | 2.37±0.54 | 0.74±0.18 |
| DD-SAM2 | Stand-Conv (3, 5) | 0.92±0.04 | 0.42±0.12 | 3.33±1.53 | 0.97±0.41 |
| DD-SAM2 | Stand-Conv (3, 7) | 0.93±0.04 | 0.43±0.10 | 2.84±1.02 | 0.83±0.28 |

Moreover, while the Dice scores of DD-Adapter were comparable to those of Stand-Conv with 3×3 and 7×7 convolutions when adapting SAM2, DD-Adapter consistently outperformed Stand-Conv in NSD, HD95, and ASD metrics. We attribute this to the significantly larger parameter count in Stand-Conv—approximately 14× more than DD-Adapter—which may increase the risk of overfitting or introduce noise into the feature representations. These results highlight the efficacy of the DD-Adapter in maintaining competitive performance while minimizing computational and parameter overhead.

### 7. Sequence Prediction on TrackRAD2025 and EchoNet Dynamic Datasets

We present the visualization results on the TrackRAD2025 dataset. In these visualizations, the ground truth annotations are shown in red, whereas the model predictions are depicted in green. This color scheme is consistently applied across all reported visualizations to facilitate comparison between predicted outcomes and reference annotations.

(1). Case A_001: s86-s90

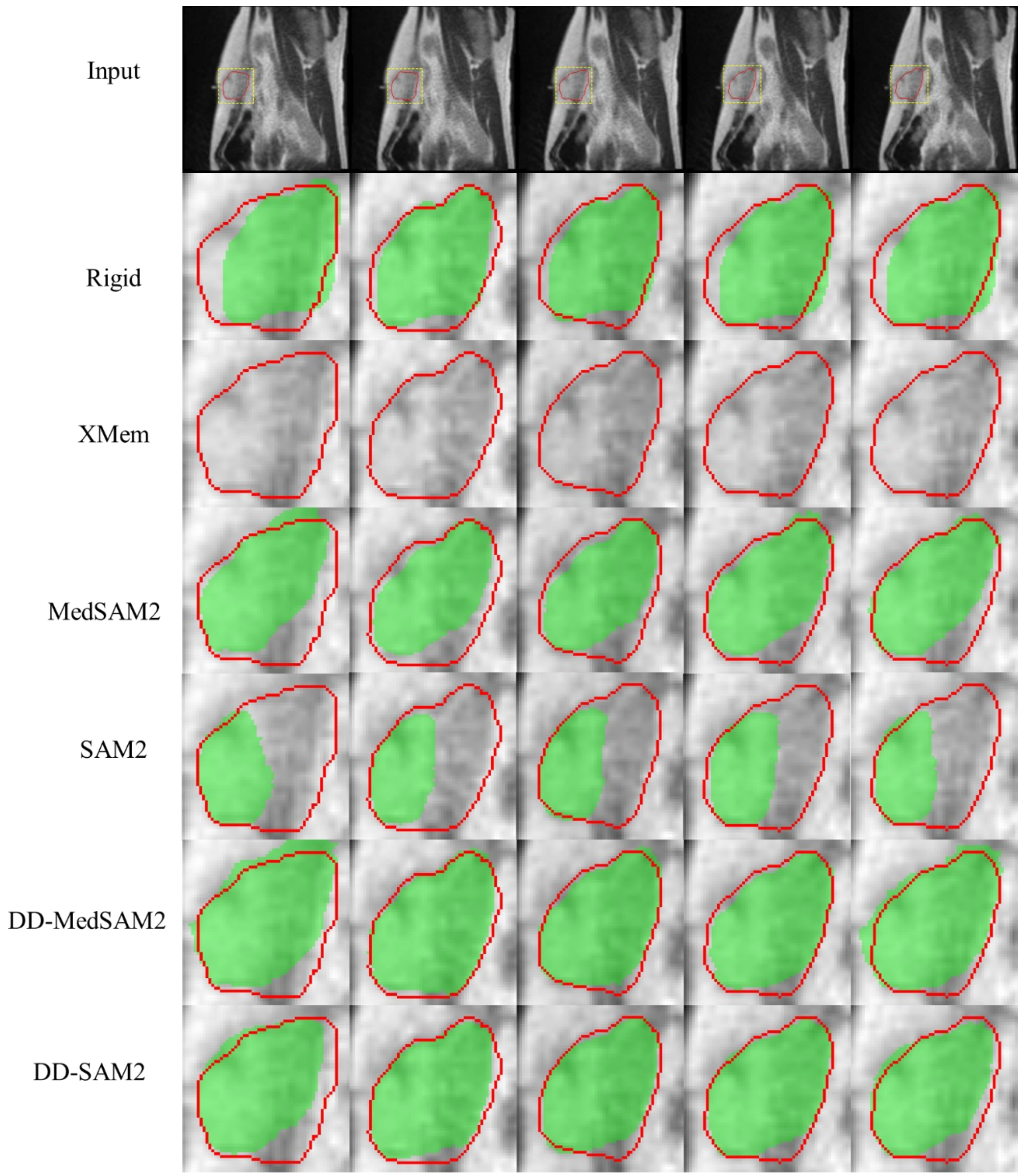

(2) Case B_012: s68-S72

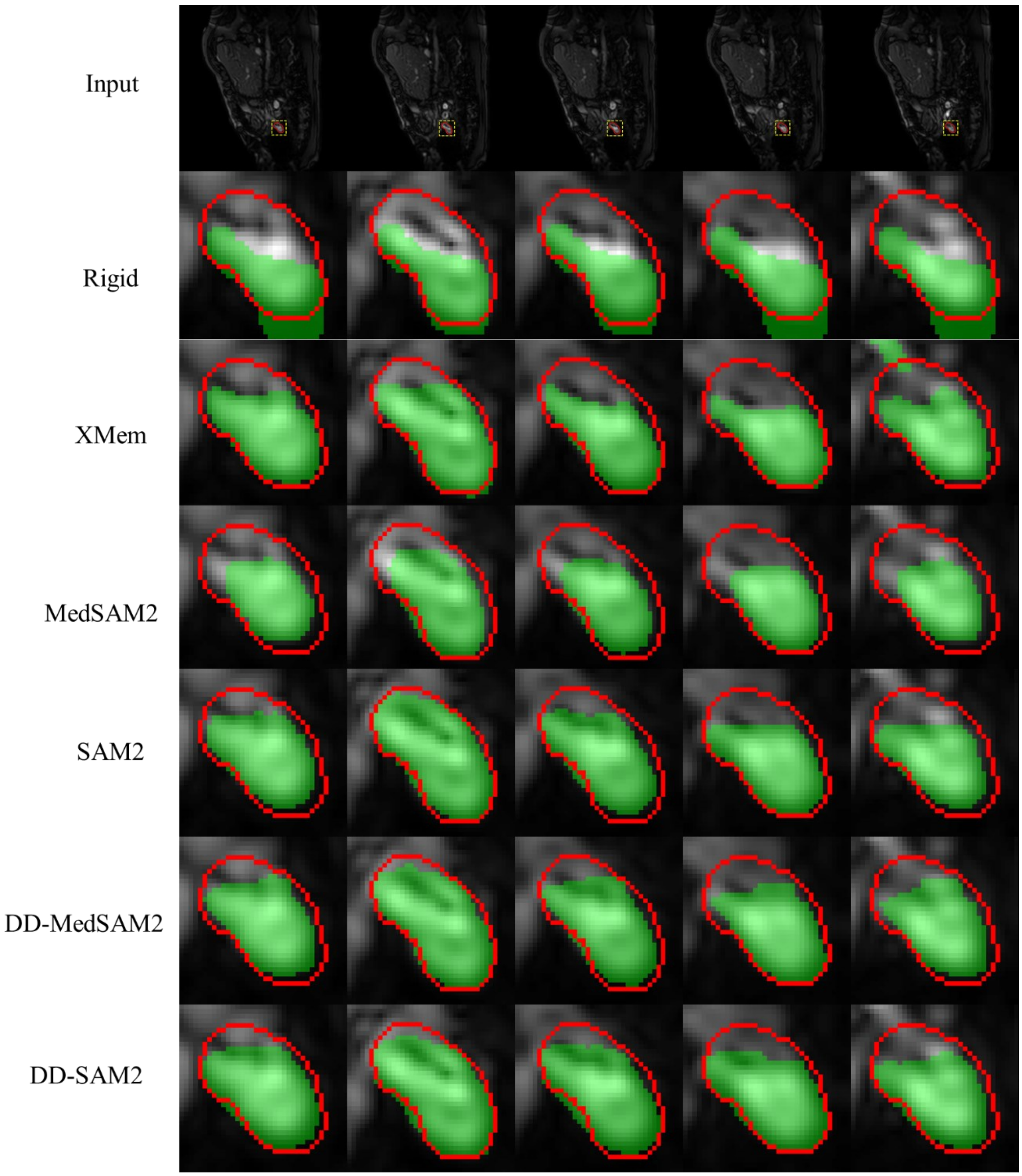

(3) Case C_008: s7-s11

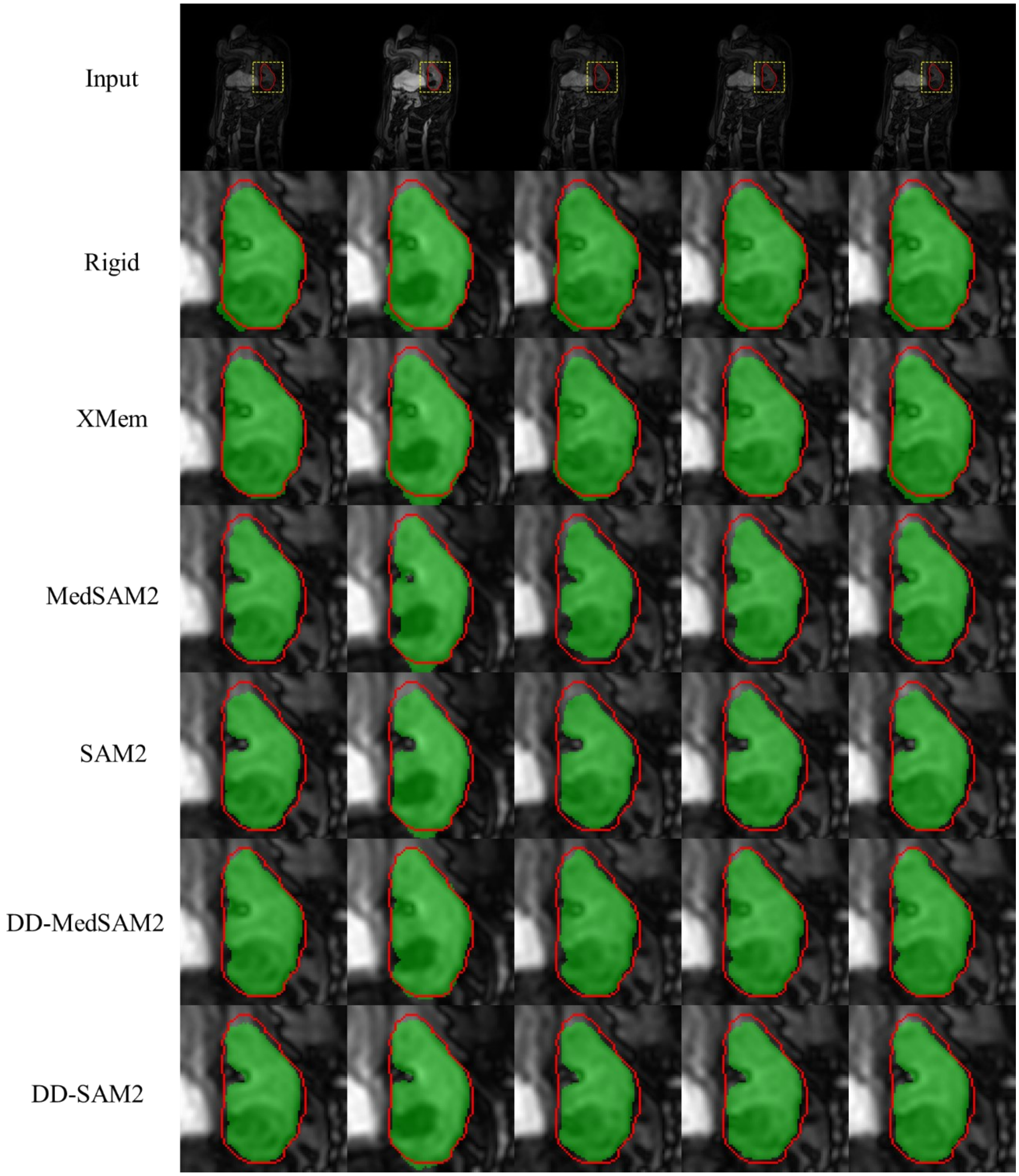
Input
Rigid
XMem
MedSAM2
SAM2
DD-MedSAM2
DD-SAM2

We further provide visualization results on the EchoNet Dynamic dataset. In these examples, the ground truth annotations are highlighted in red, while the predicted segmentations are shown in green. This consistent visualization scheme ensures clarity and facilitates direct comparison between the model predictions and reference annotations across all reported results.

(1) Case1: s176-s180

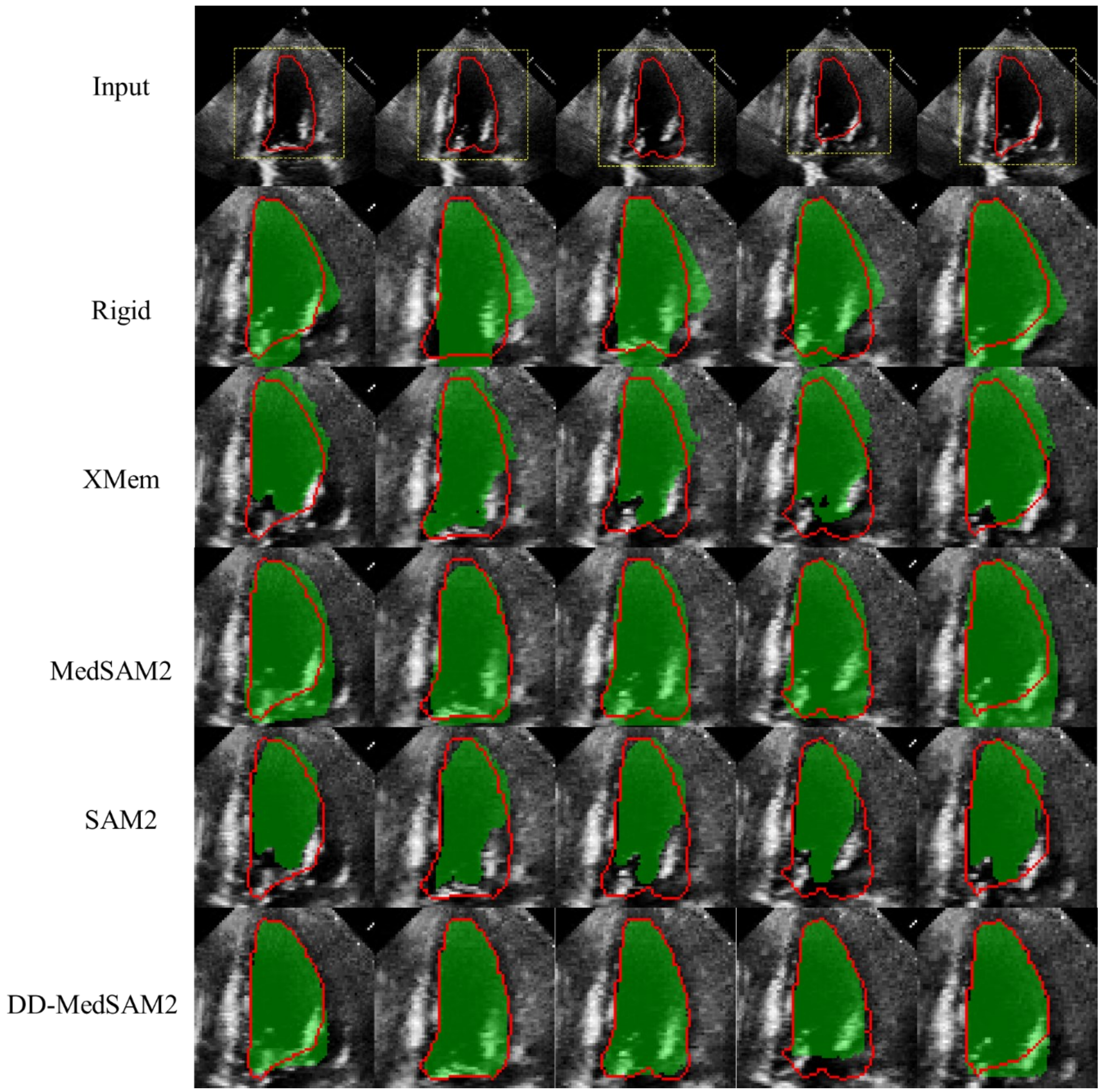

(2) Case 5: S52-s56

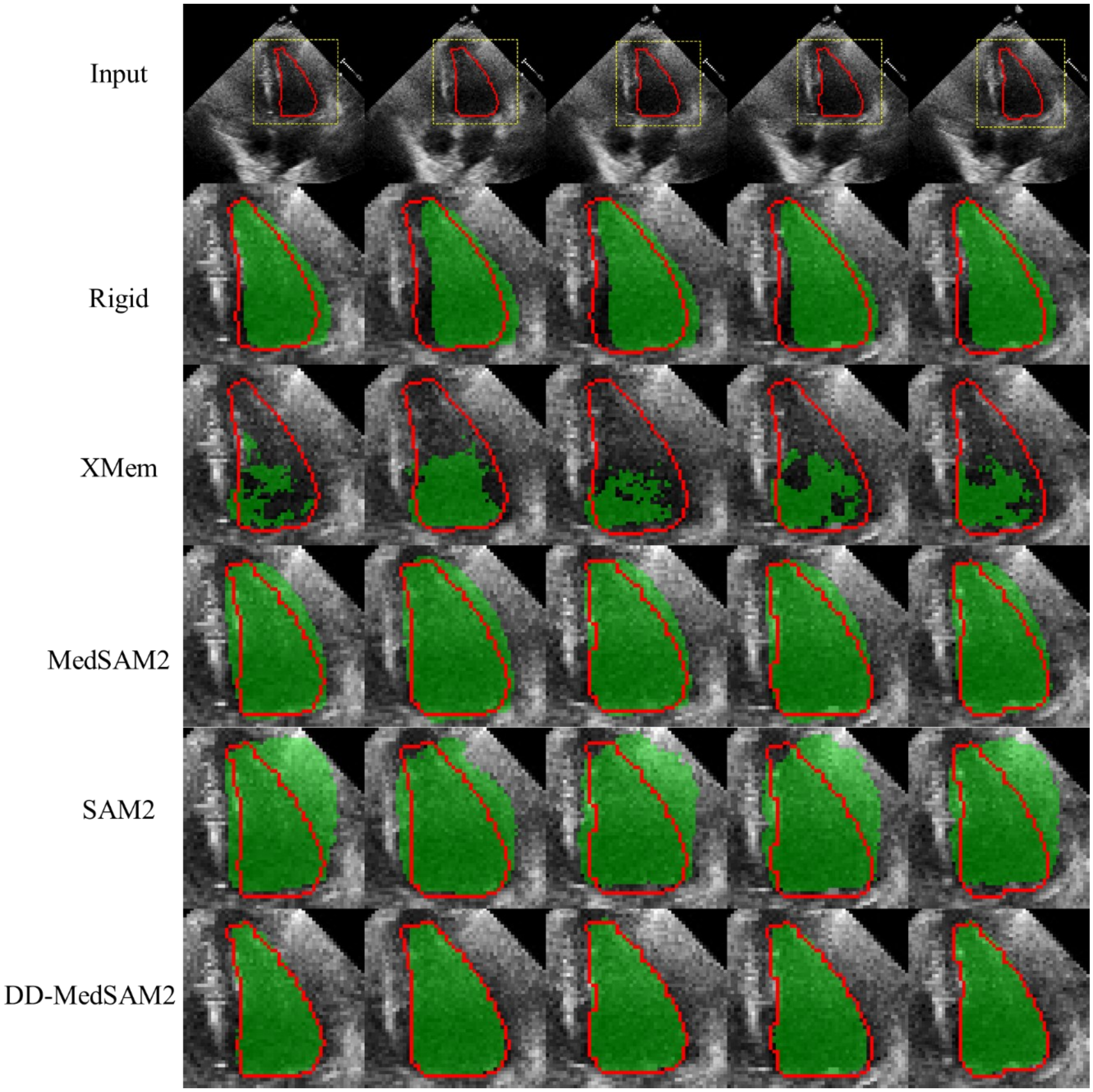

(3) Case 9: s215-s219

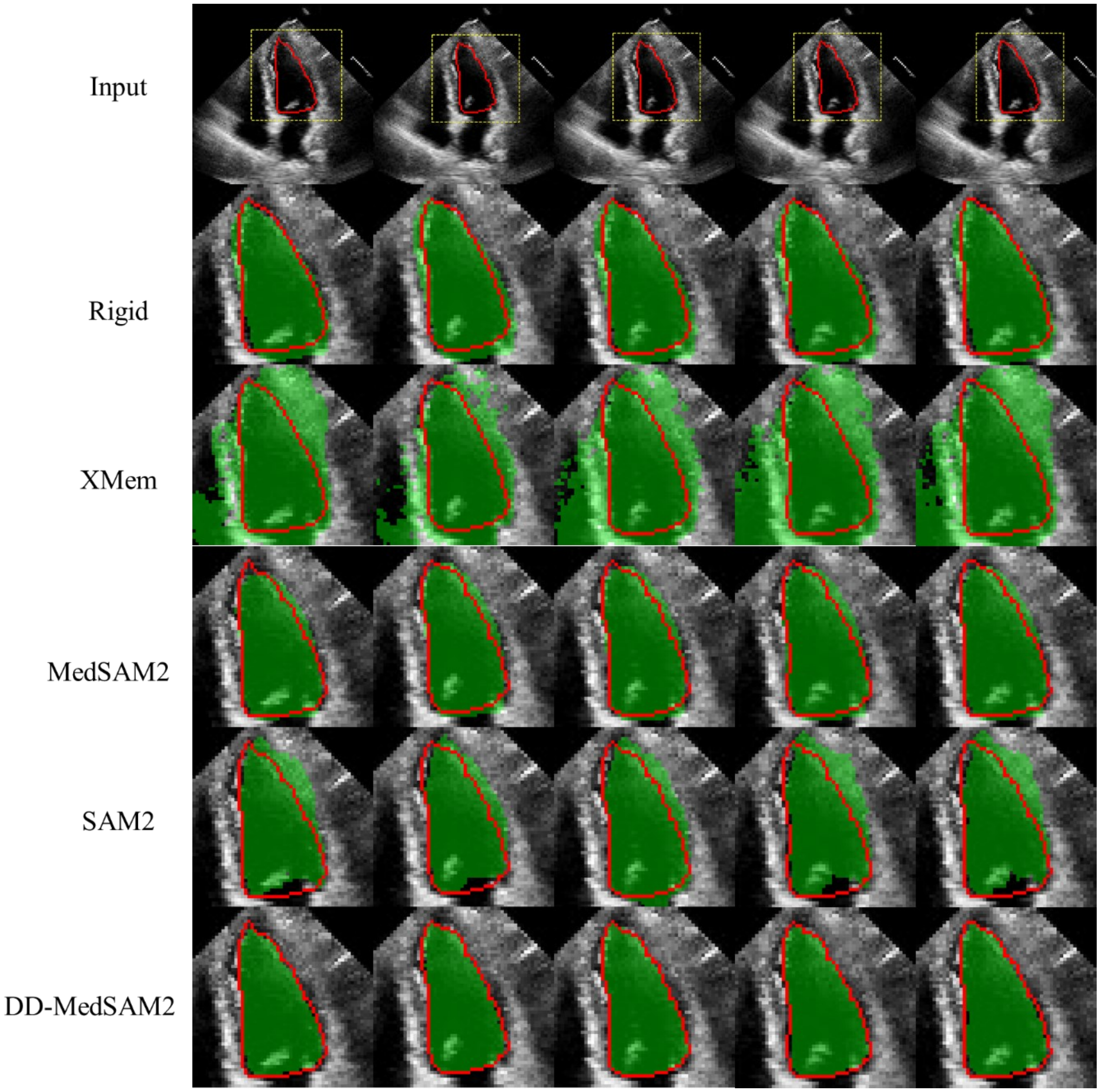